\documentclass{article}
\usepackage{multirow}

\usepackage{arxiv}

\usepackage[utf8]{inputenc} 
\usepackage[T1]{fontenc}    
\usepackage{url}            
\usepackage{booktabs}       
\usepackage{amsfonts}       
\usepackage{nicefrac}       
\usepackage{microtype}      
\usepackage{lipsum}
\usepackage{graphicx}
\usepackage{multirow}
\usepackage{orcidlink}
\usepackage{comment}
\usepackage{tcolorbox}
\usepackage{subcaption}

\usepackage[numbers]{natbib}

\definecolor{mygrey}{HTML}{949494}

\definecolor{VUB_blauw}{rgb}{0.1529, 0.2667, 0.5529}
\usepackage{hyperref}       
\hypersetup{
    colorlinks,%
    citecolor=VUB_blauw,%
    filecolor=VUB_blauw,%
    linkcolor=VUB_blauw,%
    urlcolor=VUB_blauw
}
\graphicspath{ {./images/} }
\newcommand{\customCor}[1]{%
  \includegraphics[height=1em]{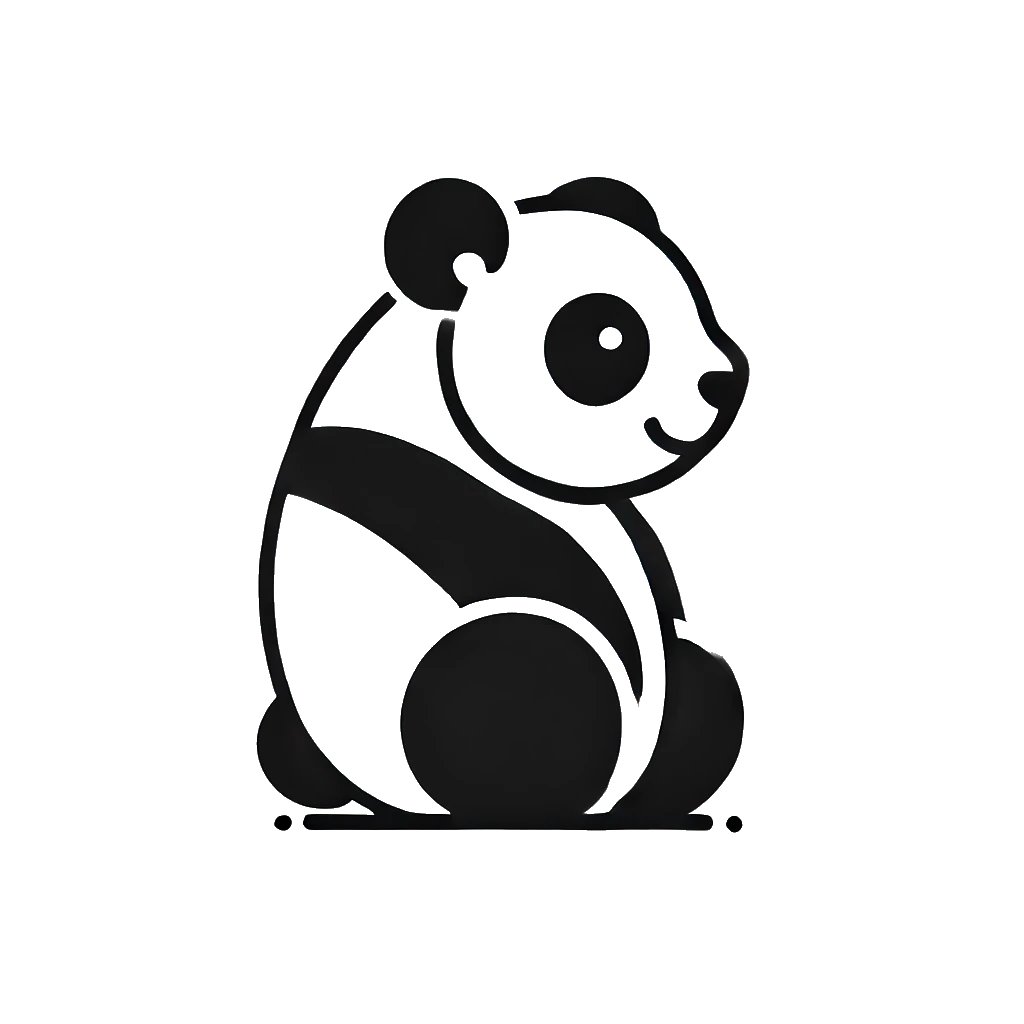} #1%
}

\AddToHook{shipout/background}{%
  \ifnum\value{page}=1 
    \pagenumbering{Roman} 
    \setcounter{page}{1} 
  \fi
  \ifnum\value{page}=2 
  \pagenumbering{arabic} 
  \setcounter{page}{2} 
  \fi
}

\title{Lexical Hints of Accuracy in LLM Reasoning Chains}
\runningtitle{Lexical Hints of Accuracy in LLM Reasoning Chains}

\author{
  Arne Vanhoyweghen\textsuperscript{1,\customCor{ }} \\ 
  \orcidlinkc{0000-0003-0103-4715} \\
  \And
  Brecht Verbeken\textsuperscript{1} \\ 
  \orcidlinkc{0000-0002-7506-3298} \\
  \And
  Andres Algaba\textsuperscript{1} \\ 
  \orcidlinkc{0000-0002-0532-3066} \\
  \And
  Vincent Ginis\textsuperscript{1,2} \\ 
  \orcidlinkc{0000-0003-0063-9608} \\
  \and
  \textsuperscript{1}Data Analytics Lab, Vrije Universiteit Brussel, 1050 Brussel, Belgium \\ 
  \textsuperscript{2}School of Engineering and Applied Sciences, Harvard University, Cambridge, Massachusetts 02138, USA
}
    
\begin{document}
\maketitle
\renewcommand{\thefootnote}{} 
\footnotetext{\includegraphics[height=1em]{panda2.png} Corresponding author: arne.vanhoyweghen@vub.be \\ Code available at: \url{https://github.com/Arne-Vanhoyweghen/CoT-R1-Sonnet-3.7-HLE-Omni-MATH} \\ Data available at: \url{https://zenodo.org/records/15648338}}
\renewcommand{\thefootnote}{\arabic{footnote}} 
\thispagestyle{plain} 

\begin{abstract}
Fine-tuning Large Language Models (LLMs) with reinforcement learning to produce an explicit Chain-of-Thought (CoT) before answering produces models that consistently raise overall performance on code, math, and general-knowledge benchmarks. However, on benchmarks where LLMs currently achieve low accuracy, such as Humanity's Last Exam (HLE), they often report high self-confidence, reflecting poor calibration. Here, we test whether measurable properties of the CoT provide reliable signals of an LLM’s internal confidence in its answers. We analyze three feature classes: (i) CoT length, (ii) intra-CoT sentiment volatility, and (iii) lexicographic hints, including hedging words. Using DeepSeek-R1 and Claude 3.7 Sonnet on both Humanity’s Last Exam (HLE), a frontier benchmark with very low accuracy, and Omni-MATH, a saturated benchmark of moderate difficulty, we find that lexical markers of uncertainty (e.g.,  \textit{guess}, \textit{stuck}, \textit{hard}) in the CoT are the strongest indicators of an incorrect response, while shifts in the CoT sentiment provide a weaker but complementary signal. CoT length is informative only on Omni-MATH, where accuracy is already high ($\approx 70\%$), and carries no signal on the harder HLE ($\approx 9\%$), indicating that CoT length predicts correctness only in the intermediate-difficulty benchmarks, i.e., inside the model's demonstrated capability, but still below saturation. Finally, we find that uncertainty indicators in the CoT are consistently more salient than high-confidence markers, making errors easier to predict than correct responses. Our findings support a lightweight post-hoc calibration signal that complements unreliable self-reported probabilities and supports safer deployment of LLMs.
\end{abstract}

\keywords{calibration errors \and chain-of-thought \and large language models \and reasoning models \and uncertainty quantification}

\section{Introduction}
Large Language Models (LLMs) can be fine-tuned with reinforcement learning to produce a Chain-of-Thought (CoT) before delivering their final response to improve their general performance~\citep{jaech2024openai,korbak2025chain,muennighoff2025s1simpletesttimescaling,wei2022chain}. The fine-tuning rewards are based both on the final correctness of the response—to improve capabilities—and on adherence to the CoT format, which may, for example, include safety-related specifications to enhance readability and alignment~\citep{guan2024deliberative,guo2025deepseek}. Although LLMs have shown a strong performance increase on demanding benchmarks, such as GPQA~\citep{rein2023gpqa}, SWE-bench~\citep{jimenez2024swebench}, and FrontierMath~\citep{glazer2024frontiermath}, they often report very high confidence while still obtaining relatively low overall accuracy, reflecting poor calibration~\citep{marjanovic2025deepseek,phan2025humanity,wei2024measuring}. This miscalibration masks silent failure modes and undermines the reliability of LLMs in open-ended settings~\citep{chen2025towards,ke2025survey}.

In addition to the poorly calibrated self-reported confidence~\citep{phan2025humanity}, the confidence of LLMs in their final responses can be assessed via sampling agreement techniques that are computationally expensive~\citep{lyu2025calibrating,Cherian2024Conformal,kadavath2022language}, or by monitoring internal representations, which requires access to the model weights~\citep{baek2025towards,templeton2024scaling,zou2023representation}. In contrast, the CoT of some LLMs is readily available with its final response, offering a potential proxy for assessing the model's internal confidence~\citep{baker2025monitoring,lindsey2025biology}. Although it has been shown that an LLM's CoT may not always faithfully reflect the model's reasoning process and may be difficult to interpret \citep{chen2025reasoning,kambhampati2025stop,korbak2025chain,stechly2025beyond}, there is some evidence that specific CoT properties predict the accuracy of the final response~\citep{jiang2025makes,wu2025phd}. For example, there is mixed evidence on how informative the length of the CoT is for predicting response accuracy~\citep{ballon2025relationship,chen2024not,illusion-of-thinking,su2025between,wang2025thoughts}.

In this paper, we investigate whether Chain-of-Thought (CoT) provides a reliable proxy for an LLM’s internal confidence by systematically analyzing three classes of features: (i) CoT length, aligning with prior work, (ii) intra-CoT sentiment volatility, and (iii) lexicographic indicators, such as hedging words. Whereas most existing studies focus narrowly on length-based correlations, our contribution is to extend this analysis with sentiment and lexical markers. We evaluate these features on two contrasting benchmarks: Omni-MATH, where state-of-the-art models already perform well, and Humanity’s Last Exam (HLE), which remains highly challenging. The results show that CoT length as well as sentiment dynamics has some predictive power in the saturated Omni-MATH setting but provides little signal on HLE, while lexical cues consistently yield informative calibration signals across both tasks.

\section{Methods}
\subsection{Datasets and Evaluation}
To analyze the lexicographic and linguistic characteristics of LLM reasoning, we collect the CoT and final responses from DeepSeek-R1~\citep{guo2025deepseek} and Claude Sonnet 3.7~\citep{anthropic2025claude37sonnet} on two large-scale benchmarks of varying difficulty: Omni-MATH~\citep{gao2024omni} and HLE~\citep{phan2025humanity}. We choose DeepSeek-R1 and Claude 3.7 Sonnet because their complete CoTs are fully available, and the serial test-time compute protocol that produces them—including the configurable maximum-token cap—is publicly documented~\citep{guo2025deepseek,anthropic2025extended}. Although the two models achieve comparable benchmark scores, Claude 3.7 Sonnet employs more extensive alignment techniques, such as constitutional fine-tuning~\citep{bai2022constitutional}, which may affect its calibration and the uncertainty signals visible in its CoT~\citep{zhu2023calibration}.

Starting from the original $3,000$-question HLE dataset~\citep{phan2025humanity}, we first remove $316$ multimodal items. From the remaining $2,684$ questions, we then exclude $412$ multiple-choice items, resulting in $2,088$ open-ended questions used in our analysis. The official benchmark results report that DeepSeek-R1 and Sonnet 3.7 achieve 8.6\% (calibration error: 81.4\%) and 8.9\% (calibration error: 88.3\%) accuracy on the full (text-only for DeepSeek-R1) version of the benchmark, respectively~\citep{cais2025hle}. Omni-MATH comprises $4,428$ mathematics problems ranging from basic arithmetic to advanced algebra and geometry. On this benchmark, reasoning models, such as OpenAI o3‑mini, achieve over $70\%$ accuracy and even surpass $80\%$ accuracy for algebra and calculus in a high reasoning effort setting~\citep{ballon2025relationship}.

Omni-MATH and HLE are intentionally contrasting benchmarks: Omni-MATH offers $4,428$ tiered math problems with automatic, high-fidelity grading~\citep{gao2024omni}, allowing us to observe CoT behavior across a continuous, well-calibrated difficulty spectrum. HLE, by contrast, poses $2,088$ multidisciplinary, open-ended questions that push models to the edge of their reasoning abilities, often revealing brittle generalization and poor calibration under genuine uncertainty~\citep{phan2025humanity}. 

For both data sets, we prompt each model with a single, standardized template designed to elicit a clear explanation, a concise final answer, and a self-reported confidence score (0\%-100\%), all the while we log its CoT:

\begin{tcolorbox}[colframe=mygrey]
Your response should be in the following format:\\

Explanation: your explanation for your final answer\\

Exact Answer: your succinct, final answer\\

Confidence: your confidence score between 0 \% and 100 \% for your answer\\

Solve the following problem: 
\end{tcolorbox}

Additionally, for Claude Sonnet 3.7, we set its max tokens to 128k and its thinking budget to 100k tokens. DeepSeek-R1 has a max tokens parameter set to 128k, which includes the CoT tokens.

To grade the final responses, we employ a customized grading strategy that accommodates the unique challenges and features of each benchmark. Omni-MATH's responses were automatically scored using Omni-Judge~\citep{gao2024omni}, an automatic mathematical grader, whose alignment with human judgment has been explicitly validated for Omni-MATH~\citep{ballon2025relationship,gao2024omni,verga2024replacing}. 

We prompted Omni-Judge as follows:
\begin{tcolorbox}[colframe=mygrey]
\textbf{OBJECTIVE} \\

You are tasked with evaluating the correctness of a student's answer. Below, you are provided with a problem, a reference answer, and a student's answer. You should assess whether the student's answer captures the same meaning as the reference answer, even when expressed with different wording or format. \\

Your tasks include: \\
A. Identify Mathematical or Notational Equivalence. \\
B. Conclude with a brief explanation as to why the student's output is correct or incorrect. \\

\textbf{\# RESPONSE: MARKDOWN REPORT \#} \\
\textbf{Student Final Answer} \\
Extract the student's final answer, which is enclosed in ``\textbackslash boxed\{\}''. \\

\textbf{Equivalence Judgement} \\
Whether the student's answer shares the same meaning with the reference answer. (TRUE or FALSE) \\

\textbf{Justification} \\
Conclude with a brief explanation as to why the student's answer is correct or incorrect. \\

\textbf{ATTENTION} \\
- The reference answer is ALWAYS correct. You should carefully judge whether the student gives the same answer as the reference answer. \\
- The answer is FALSE even if the student's final answer is almost correct with a minor mistake. \\
- The answer is contained within the ``boxed'' section, so you can focus solely on comparing the content in the student's answer box with the reference answer, without needing to consider the intermediate steps. \\
- Add ``=== report over ==='' at the end of the report.
\end{tcolorbox}

For HLE, each answer is independently assessed by both a human expert and a different LLM. We use OpenAI o3‑mini to avoid preference leakage, which can occur when the same model is used for both answer generation and evaluation, potentially resulting in more lenient or biased assessments of its own responses~\citep{li2025preference}. The agreement between human and automated grades is high: 93.9\% for Claude 3.7 Sonnet and 93.4\% for DeepSeek-R1, corresponding to Cohen's kappa values of 0.88 and 0.87, indicating near-perfect alignment. For further analysis, we include only HLE items where human and OpenAI o3‑mini evaluations agree. Finally, to assess model calibration, we group the outputs into decile confidence intervals (0\% - 10\%, 10\% - 20\%, etc.) and calculate the mean absolute difference between the average reported confidence of each bin and its empirical precision.

Both DeepSeek-R1 and Claude Sonnet 3.7 perform similarly across the two benchmarks and in line with their official benchmark results. Note that a small derivation is to be expected as Bowyer et al. (2025)~\citep{bowyer2025position} argue, point estimates on small, specialized tests should be reported with confidence intervals to avoid underestimating uncertainty (e.g., error bars or significance tests). On HLE, Claude Sonnet 3.7 achieves a slightly higher accuracy than DeepSeek-R1 (9.2\% vs. 8.9\%), whereas on Omni-MATH, DeepSeek-R1 outperforms Claude Sonnet 3.7 (72.5\% vs. 69.1\%). In both cases, Claude Sonnet 3.7 displays the highest calibration error. The large difference in accuracy between Omni-MATH and HLE highlights the high complexity of HLE. A global overview of accuracies and calibration errors is presented in Table~\ref{accuracies}. A full breakdown, categorized by difficulty level (only for Omni-MATH), is provided in Appendix~\ref{Accuracy Breakdown}.

\begin{table}[ht]
\centering
\resizebox{0.75\textwidth}{!}{%
\begin{tabular}{lcccc}
\midrule
\textbf{Dataset}                    & \textbf{Model}             & \textbf{Accuracy (\%)} & \textbf{Calibration Error (\%)} &  \\ \cline{1-5}\\
\multirow{2}{*}{HLE}       & DeepSeek-R1       & 8.6\%         & 78.1\%                 &  \\ 
                           & Claude 3.7 Sonnet & 9.2\%            & 84.6\%                     &  \\ \\ \cline{1-5}\\
\multirow{2}{*}{Omni-MATH} & DeepSeek-R1       & 72.5\%        & 20.1\%                 &  \\ 
                           & Claude 3.7 Sonnet & 69.1\%        & 29.0\%                 &  \\ \\ \midrule\\
\end{tabular}%
}
\caption{Overall accuracy and calibration error for DeepSeek-R1 and Claude 3.7 Sonnet on the HLE and Omni-MATH benchmarks. While both models perform similarly, Claude Sonnet 3.7 is slightly more accurate on HLE, and DeepSeek-R1 outperforms Claude on Omni-MATH. Claude consistently exhibits higher calibration error across both benchmarks.}
\label{accuracies}
\end{table}

\subsection{intra-CoT sentiment Volatility in the Chain-of-Thought}
The ability to regulate emotions, such as maintaining composure or overcoming moments of doubt, has been linked to improved cognitive outcomes in humans~\citep{schutz2004emotional,schutz2000emotions,schutz2008approach,harley2021university}. Motivated by this, we examine the impact of the Intra-CoT sentiment volatility within the CoT responses on the response correctness. Specifically, we asked whether the model's reasoning began with a confident, optimistic tone and ended in doubt, or if it started hesitantly and grew more optimistic as the answer developed.

To quantify this, we process each CoT through OpenAI o3‑mini as an independent sentiment evaluator~\citep{tori2024performing}. For each rationale, we extract two sentiment scores—the most negative and the most positive sentiment, ranging from -1 (strongly negative) to +1 (strongly positive). We also record the order in which these sentiments appear in the CoT, i.e., whether the response moved from negative to positive (an ``upward'' direction, scored as 1), from positive to negative (``downward,'' scored as –1), or remained neutral throughout (scored as 0). We use the following prompt to evaluate each CoT:

\begin{tcolorbox}[colframe=mygrey]
Judge how emotional this text is by reporting on its most positive emotion and its most negative emotion on a scale from -1 (negative) to 1 (positive), and the order in which they appear, i.e. direction in which they appear, if the most negative emotion appears before the most positive emotion the direction is upward so, 1, if inverse is true the direction is -1 if the text is completely neutral and no gradient exists present between the two emotional extremes the direction is 0, picking up on small effects. Always end your response with: \\
\\
Most positive emotion Score: ... \\
Most negative emotion Score: ... \\
Direction: ...
\end{tcolorbox}

Using the sentiment scores provided by OpenAI o3‑mini, we define Intra-CoT sentiment volatility as the signed difference between the most positive and most negative sentiment within each CoT. Large differences ($\Delta$) may indicate moments of uncertainty, backtracking, or sudden shifts in confidence, while smaller deltas suggest a more consistent and steady tone throughout the reasoning process. 

\subsection{Lexicographic Analysis of the Chain-of-Thought}
To explore the connection between the linguistic characteristics of model-generated rationales and their accuracy, we perform a lexicographic analysis of CoT responses from all evaluated datasets and models. For each unique word found within a CoT, we calculate its relative accuracy by comparing the success rate of responses containing that word to the average accuracy within the respective dataset. To ensure reliability, we exclude low-frequency words by requiring that each word appear at least a minimum number of times (300 occurrences) in every dataset to be included in the analysis.

\subsubsection{Hedging Words and Accuracy}
In addition to the general lexicographic analysis, we focus on a specific subset of the lexicon: hedging words. In human communication, hedging—through modal verbs (e.g., might, could), uncertainty adverbs (e.g., possibly, perhaps), and tentative phrases (e.g., it seems that)—serves as a cue indicating lower confidence ~\citep{hyland1998boosting,demir2018hedging,lakoff1973hedges}. We hypothesize that the same might be true for the CoT of LLMs. 

Drawing on established taxonomies of uncertainty markers,~\citep{hyland1998boosting,demir2018hedging,lakoff1973hedges}, we compiled a lexicon of such expressions and computed a hedging rate per, i.e., proportion of sentences in a rationale containing at least one hedging expression CoT. The full hedging lexicon used was:

\begin{itemize}
    \item Modal and uncertainty verbs: might, may, could, should, would, seems, suggests, appears
    \item Uncertainty adverbs: possibly, perhaps, likely, unlikely, probably, generally, usually, sometimes, often, tends, somewhat, rather, quite, almost, nearly, virtually, presumably, arguably, relatively, fairly, reasonably, mostly, partially, mainly, primarily, essentially, basically
    \item Common hedging phrases: it seems that, it appears that, it suggests that, it is possible that, it is likely that
    \item Additional qualifiers: in part, to some extent
\end{itemize}

\section{Results}
\subsection{Relationship Between CoT Length and Accuracy}
\begin{figure}[t]
    \centering
    \begin{subfigure}[t]{0.48\linewidth}
        \centering
        \includegraphics[width=\linewidth]{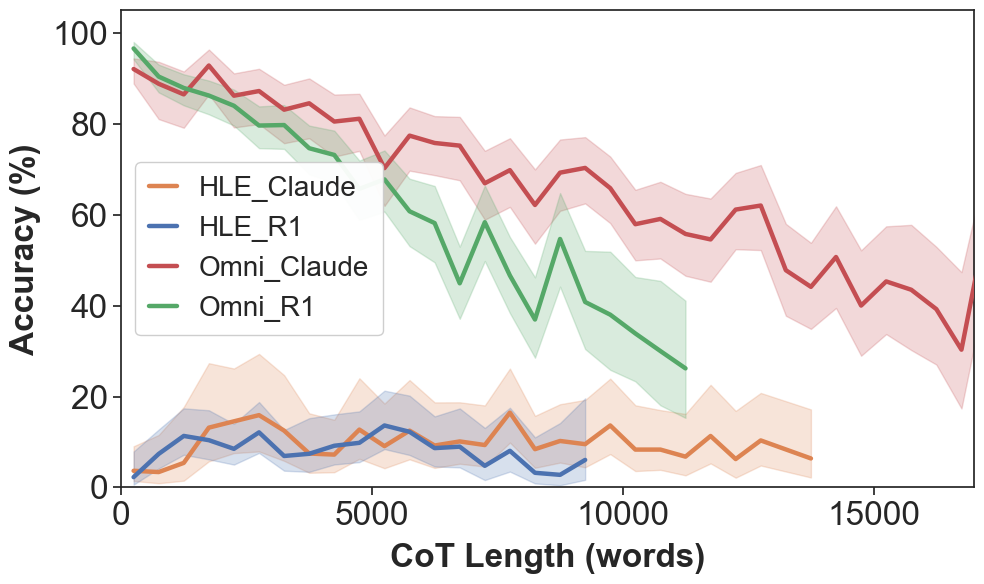}
        \label{reasoning accuracy}
    \end{subfigure}
    \hfill
    \begin{subfigure}[t]{0.48\linewidth}
        \centering
        \includegraphics[width=\linewidth]{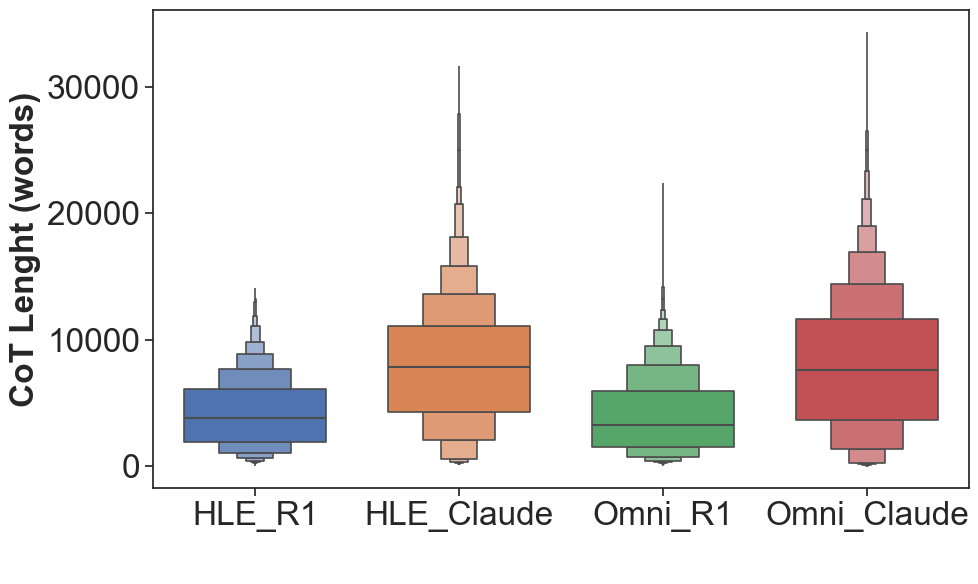}
        \label{reasoning length}
    \end{subfigure}
    \caption{Accuracy as a function of CoT length (left) and the distribution of CoT lengths (right) for each model and benchmark. On Omni-MATH, accuracy declines with longer responses, especially for DeepSeek-R1; in HLE, accuracy remains low and variable regardless of length. CoT length bins with fewer than 30 samples were excluded from the accuracy analysis on the left. Claude Sonnet 3.7 generally produces longer chains-of-thought than DeepSeek-R1, likely a reflection of different reasoning allowances.}
    \label{fig:reasoning-accuracy-combined}
\end{figure}

To assess how the length of the CoT relates to response correctness, we analyze accuracy as a function of the number of words in each CoT rationale. We observe a downward trend in the Omni-MATH benchmark: accuracy declines as CoT length increases (see Figure~\ref{fig:reasoning-accuracy-combined}). Claude 3.7 Sonnet proves more robust to longer reasoning chains than DeepSeek-R1, losing on average 3\% of accuracy per additional 1,000 CoT words, compared to 6.2\% for DeepSeek-R1. These results echo the findings of~\citep{ballon2025relationship}, who report that o3-mini-high loses about 0.8 \% per 1,000 additional reasoning tokens on Omni-MATH benchmark. By contrast, in HLE, we find that the CoT length shows no discernible pattern, accuracy stays low, and fluctuates widely regardless of CoT length. This suggests that CoT length predicts correctness only for benchmarks of intermediate difficulty—that is, tasks within the model’s demonstrated capabilities but not yet saturated. These findings are consistent with previous work showing a negative relationship between CoT length and response accuracy on Omni-MATH~\citep{ballon2025relationship}, as well as the disappearance of this effect on the most challenging benchmarks~\citep{illusion-of-thinking,opus2025comment}. In general, we find that Claude Sonnet 3.7 generally produces longer CoTs than DeepSeek-R1, which may be due to architectural choices or differences in reinforcement learning post-training~\citep{muennighoff2025s1simpletesttimescaling,guo2025deepseek}.

\subsection{Intra-CoT sentiment Volatility and Reasoning Dynamics}
\begin{figure}[t]
    \centering
    \begin{subfigure}[t]{0.48\linewidth}
        \centering
        \includegraphics[width=\linewidth]{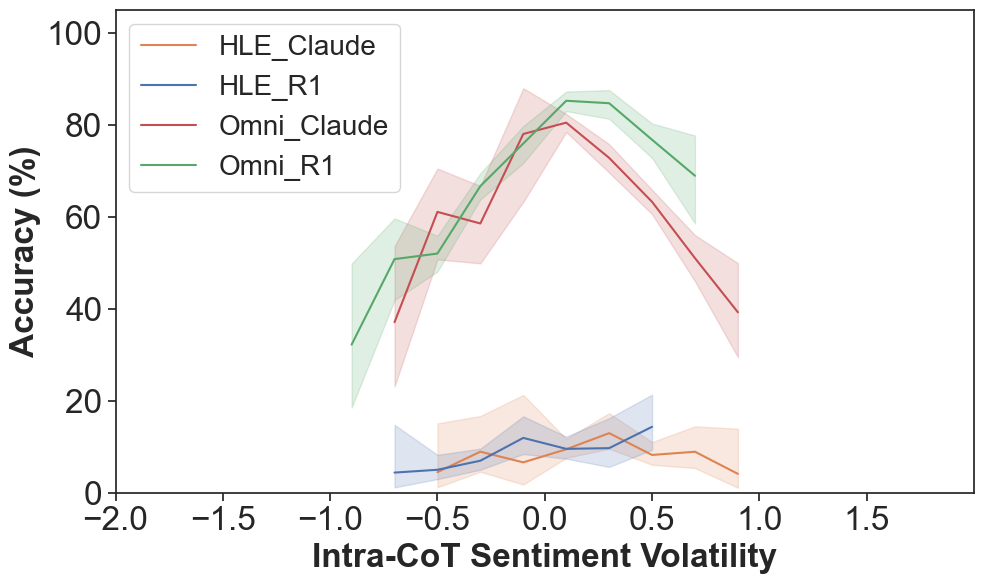}
        \label{emotion accuracy}
    \end{subfigure}
    \hfill
    \begin{subfigure}[t]{0.48\linewidth}
        \centering
        \includegraphics[width=\linewidth]{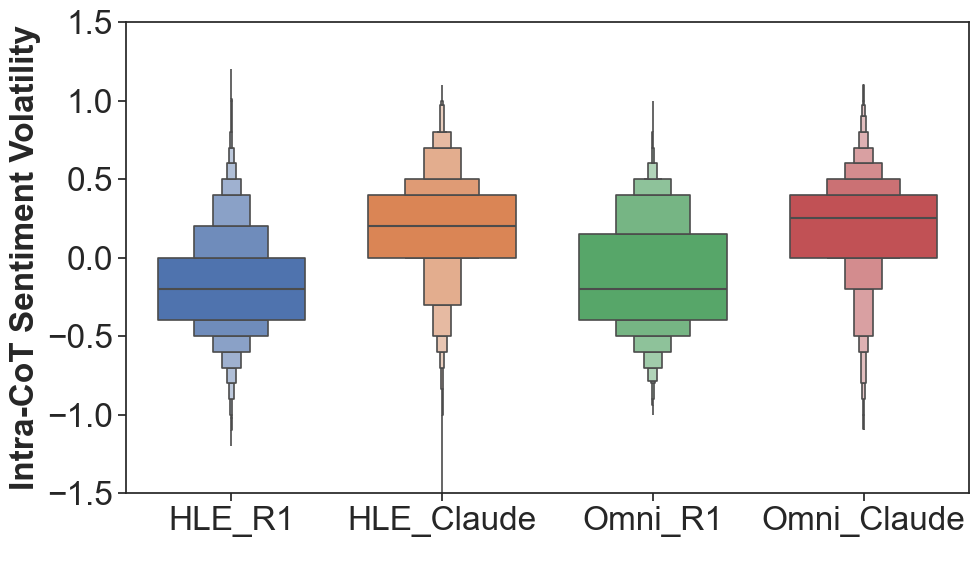}
        \label{emotion}
    \end{subfigure}
    \caption{Accuracy as a function of Intra-CoT sentiment volatility (left) and its distribution (right) for each model and benchmark. On Omni-MATH, accuracy peaks for CoTs with low or slightly positive volatility; for HLE, accuracy is uniformly low. Claude 3.7 Sonnet generally produces more positive-toned reasoning, while DeepSeek-R1 leans more to a slightly negative flow of emotion. Bins with fewer than 30 samples were excluded from the left figure.}
    \label{fig:emotion-hedging-combined}
\end{figure}

Figure~\ref{fig:emotion-hedging-combined} shows the relationship between Intra-CoT sentiment volatility within CoT and response correctness. We observe that a parabolic pattern emerges for Omni-MATH, with accuracy peaking at a volatility value of approximately 0.1. This suggests that CoTs with a consistent or slightly uplifting sentiment are associated with the highest accuracy. In contrast, no such pattern is observed for HLE, where accuracy remains uniformly low regardless of the degree of sentimental variation. To rule out the possibility that a nonzero mean (baseline) emotion score was driving this effect, we repeated the volatility‐accuracy analysis on mean‐centered sentiment values. After centering, we found the same parabolic trend for Omni-MATH and the same flat pattern for HLE. While both reasoning models display a predominantly neutral CoT, they exhibit markedly different sentimental footprints. Claude 3.7 Sonnet's responses tend to be slightly optimistic in tone, becoming more positive over the course of their reasoning, while DeepSeek's rationales often exhibit a slight negative shift. This is marked by the more positive Intra-CoT sentiment volatility of Claude 3.7 Sonnet and DeepSeek-R1, which has more weight on the negative side in Figure~\ref{fig:emotion-hedging-combined}(right).

\subsection{Lexical Hints of Uncertainty}
To scrutinize how specific words reflect or undermine a model's underlying confidence in response correctness, we construct a lemmatized lexicon from the CoT. We exclude rare or highly exclusive words from the lemmatization process, retaining only those appearing in at least 300 CoT responses for each model and benchmark. We distinguish between lemmatized words that consistently harm or boost accuracy. For ease of reading, we use words to refer to the lemmas in the remainder of the paper.

\begin{figure}[t]
    \centering
    \includegraphics[width=1\linewidth]{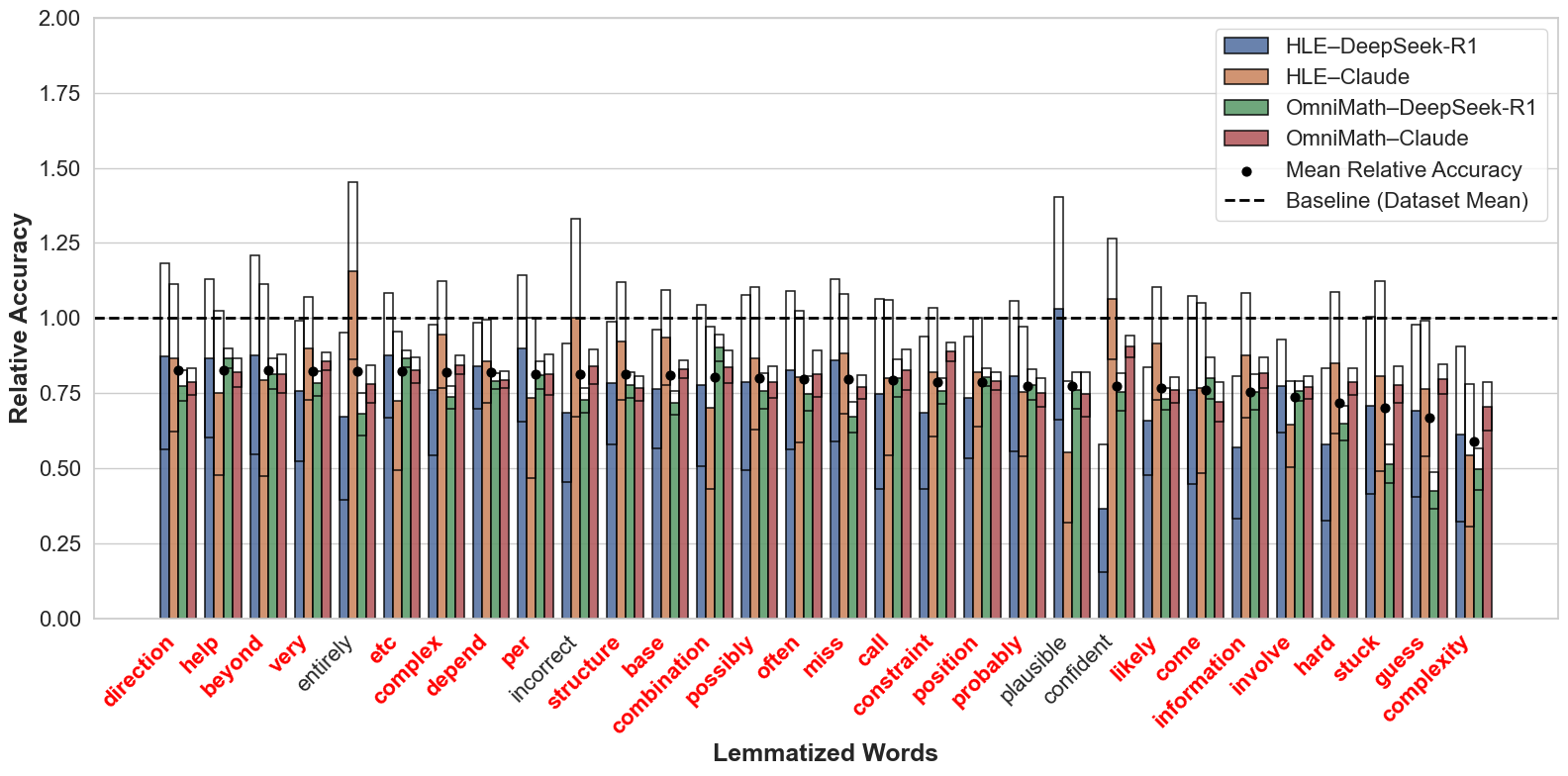}
    \caption{The relative accuracy per dataset/model is shown as a barplot on which words that consistently reduce accuracy have been highlighted in red. Moreover, the dataset mean is marked by a dashed line, and the mean relative accuracies by black dots. Remarkably, many of the words most strongly associated with reduced accuracy are typically linked to uncertainty, cognitive overload, or confusion (e.g., complexity, guess, stuck, hard, likely, possibly, depend, help, direction). These patterns mirror human language for expressing task difficulty and suggest linguistic convergence between models and humans.}
    \label{Dragging Words}
\end{figure}

Interestingly, many of the words most strongly linked to reduced accuracy, such as \textit{complexity}, \textit{guess}, \textit{stuck}, \textit{hard}, \textit{likely}, \textit{probably}, \textit{possibly}, and \textit{complex}, are intuitively associated with task difficulty and cognitive overload, mirroring language that humans use to express uncertainty or task difficulty~\citep{renkl1997learning,hyland1998boosting,demir2018hedging,lakoff1973hedges}. This suggests a degree of linguistic convergence in the signaling of uncertainty between reasoning models and humans. As shown in Figure~\ref{Dragging Words}, the top 30 words most detrimental to accuracy also include terms conveying confusion or lack of direction, such as \textit{information}, \textit{miss}, \textit{depend}, \textit{beyond}, \textit{help}, and \textit{direction}. A full overview of these lexical hints of uncertainty, i.e., ``harmful'' words can be found in Appendix~\ref{appendix: All Consistently Harmful Words}.

In contrast, only the word ``so'' appears to have consistently boosted accuracy across all datasets (see Figure~\ref{Booster Words}). However, when considering its confidence interval, this apparent effect is likely due to random variation rather than a true underlying pattern. As a general rule, the words with the highest mean relative accuracy tend to be those characteristic of mathematical reasoning, including \textit{equivalent}, \textit{solve}, \textit{therefore}, \textit{substitute}, \textit{equal}, \textit{equation}, \textit{calculation}, \textit{formula}, \textit{compute}, and \textit{coefficient}. This is an artefact of both Claude and DeepSeek demonstrating above-average performance on mathematical questions within HLE, achieving accuracies of 10\% and 9.1\%, respectively.

\subsubsection{Hedging}
We continue by looking at a subset of our full lexicon, hedging words. Hedging words are typically used by humans to mark doubt or a lack of knowledge~\citep{hyland1998boosting,demir2018hedging,lakoff1973hedges}; the same could be true for LLM-generated CoT. 

\begin{figure}[t]
    \centering
    \begin{subfigure}[t]{0.48\linewidth}
        \centering
        \includegraphics[width=\linewidth]{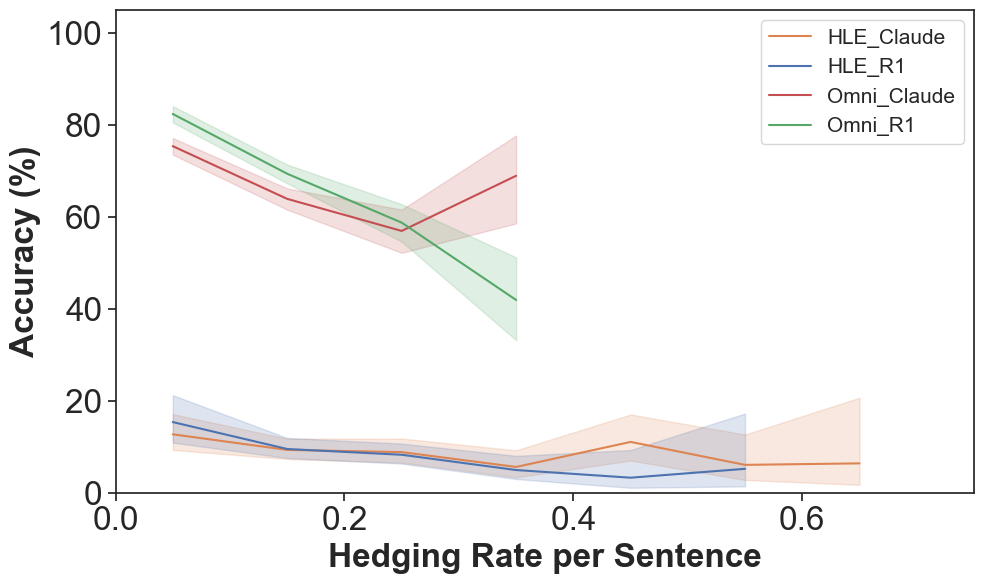}
        \label{Hedging Accuracy}
    \end{subfigure}
    \hfill
    \begin{subfigure}[t]{0.48\linewidth}
        \centering
        \includegraphics[width=\linewidth]{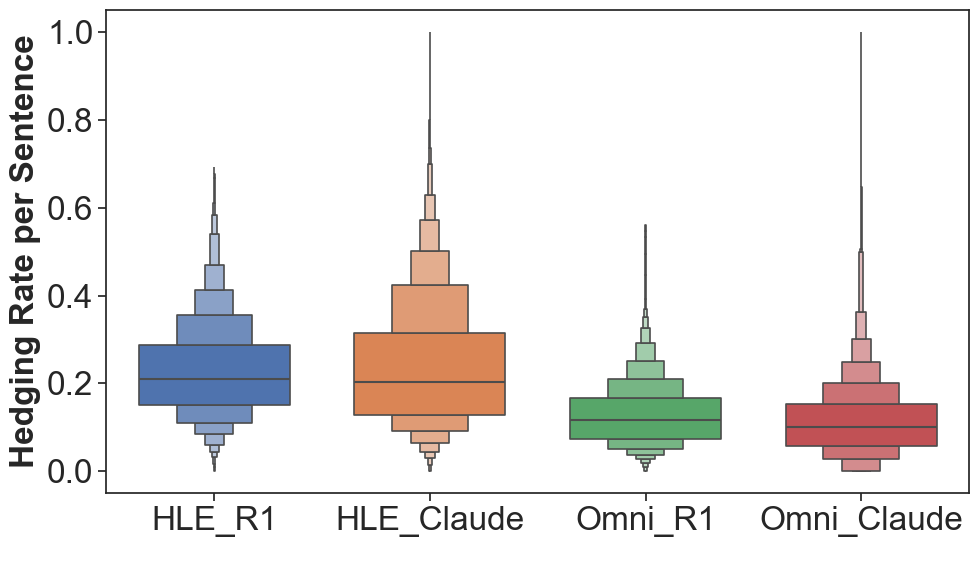}
        \label{Hedging Boxenplot}
    \end{subfigure}
    \caption{Accuracy as a function of hedging rate (left) and the distribution of hedging rates (right) for each model and benchmark. Higher hedging rates are associated with lower accuracy, especially on Omni-MATH, though the effect is small. Hedging is more common in HLE, with a median rate of 20\%, compared to 10\% for Omni-MATH, reflecting increased linguistic uncertainty on more challenging tasks. As before, bins with fewer than 30 samples were excluded from the left figure.}
    \label{Hedging}
\end{figure}

Across the four model benchmark pairs, we observe a negative relationship between hedging rate and accuracy: as the proportion of hedged sentences increases, the accuracy tends to decline (Figure~\ref{Hedging}). This trend is most pronounced for Omni-MATH, where the correlations are $r=-0.24$ $(p<0.001)$ for DeepSeek and $r=-0.14$ $(p<0.001)$ for Claude 3.7 Sonnet, indicating small but statistically detectable negative associations under the null hypothesis of zero correlation. For HLE, the correlations are weaker: $r=-0.10$ $(p<0.001)$ for DeepSeek-R1, indicating a statistically significant deviation from zero, and $r=-0.04$ $(p=0.06)$ for Claude 3.7 Sonnet, which does not differ significantly from zero at conventional thresholds. Overall, while the negative association is consistent in direction, the effect sizes are small.

Moreover, we observe that hedging is more prevalent when models address the more challenging HLE benchmark. In Omni-MATH, the median hedging rate is about 10\%, which means that roughly one in ten sentences contains a hedging expression. For HLE, this median rises to about 20\%. This pattern, shown in Figure~\ref{Hedging}, suggests that as task difficulty increases, models produce more linguistic signals of uncertainty in their reasoning.

\subsection{Prediction Accuracy}
The question now remains whether these lexicographic features can be used to predict/anticipate the correctness of a reasoning model's final response. To assess which features best predict the correctness of model-generated reasoning, we train a neural network on both general features (CoT length, emotion volatility, and hedging rate) and with the 25 most consistently ``harmful'' non-lemmatized words (see Appendix~\ref{tab:harmful-words}, and combinations between the standard features and ``harmful'' words. We then test both its in-benchmark and cross-benchmark capability, by testing its performance on the benchmark it is trained on and the other benchmark.

For all four experiments, we employ a feed-forward neural network with two hidden layers (32 and 16 units, ReLU activations) and a sigmoid output. Models are trained using the Adam optimizer and binary cross-entropy loss. Given the strong class imbalance in the training data, particularly for HLE, whose accuracy for the minority class was only 8.9\%, it is crucial to prevent the model from exploiting this imbalance with trivial predictions (e.g., always guessing ``incorrect'' for DeepSeek's answers on HLE would yield an accuracy of 91.4\%). Consequently, we apply class weighting during neural network training, penalizing mistakes on the minority class more heavily, to encourage the model to learn non-trivial, meaningful patterns for both classes. All experiments are conducted on 30 seeds (0-29) to ensure robust estimation, with the results averaged across multiple runs. Training is performed on 80\% of the original unbalanced data; we then balance the remaining 20\% so that evaluation sets contain the same number of correct and incorrect examples, ensuring that performance consistently above 50\% is indeed better than random guessing.

\begin{figure}[t]
    \centering
    \includegraphics[width=1\linewidth]{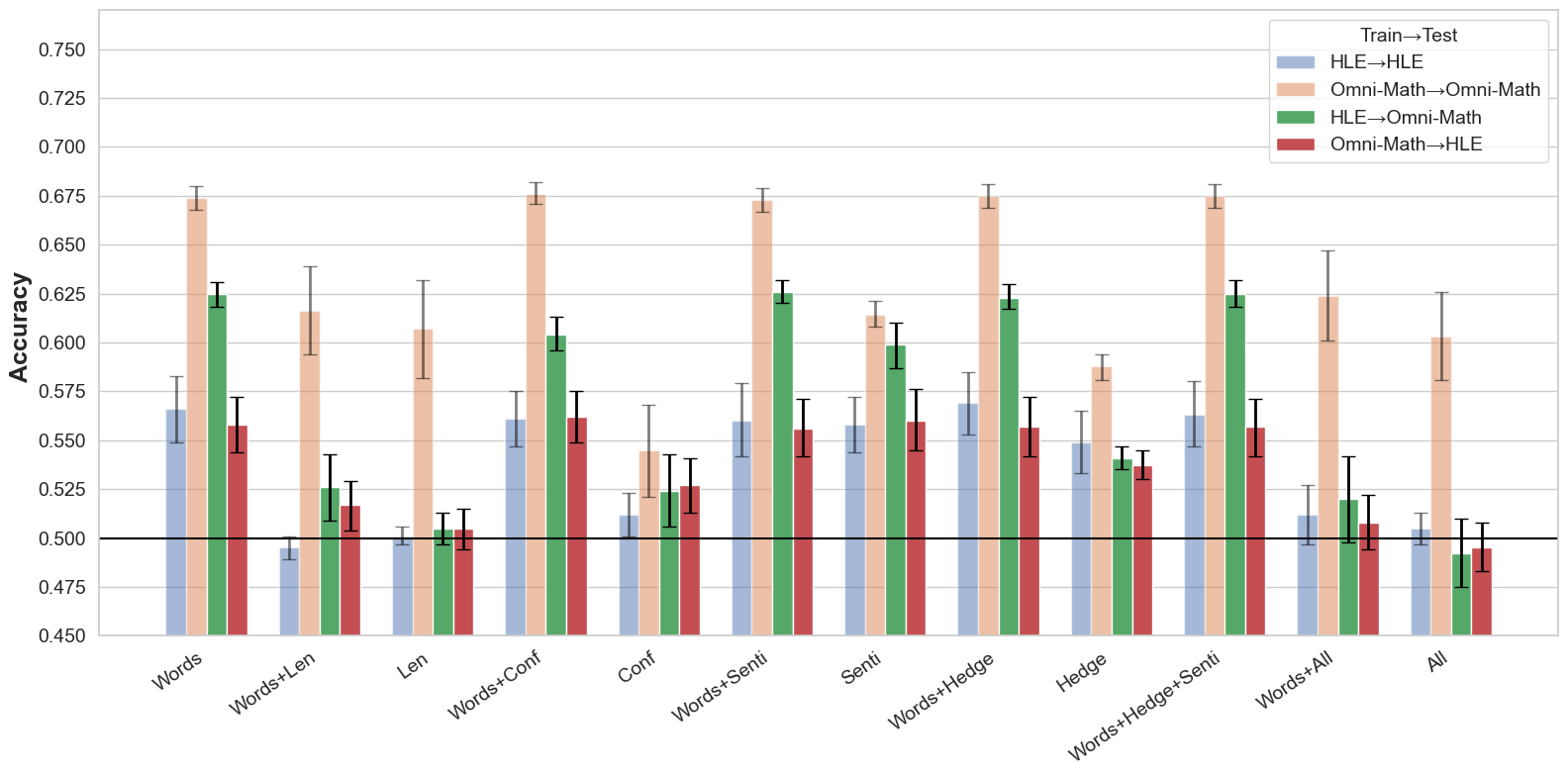}
    \caption{Each group of bars represents a different combination of input features: bag-of-words using 25 “harmful” words (“Words”), common surface features (CoT length, emotion, hedging), or their combinations. Neural nets were always trained on imbalanced data but evaluated on balanced test sets to ensure accuracy reflects true generalization rather than label bias. Results are averaged across 30 random seeds, with error bars showing variability. In-benchmark (blue, orange) performance is generally higher than out-of-benchmark (green, red). Adding additional emotion and hedging did not significantly improve cross-benchmark generalization, while adding length harmed both in-benchmark and cross-benchmark accuracy. This suggests that a small, targeted lexicon of harmful words serves as the most generalizable predictor of answer correctness across benchmarks.}
    \label{NN}
\end{figure}

Figure~\ref{NN} illustrates that adding additional features, such as CoT length or hedging rate, do not significantly enhance performance and, in some cases, even reduce cross-benchmark accuracy. This pattern is consistent across all random seeds and experimental setups. Notably, CoT length, which has been highlighted in previous studies~\citep{ballon2025relationship,chen2024not,illusion-of-thinking,su2025between,wang2025thoughts}, consistently provides the least predictive value and harms performance when included as an additional feature in all of our analyses. The ROC curves for each analysis can be found in Appendix~\ref{ROC}. These results suggest that this small set of words offers the most generalizable signal for predicting final response correctness, serving as lexical hints of uncertainty within CoT reasoning. The complete confusion matrices and Matthews Correlation Coefficient (MCC) for the best-performing neural network, which was trained using only these 25 words, are available in Appendix~\ref{confusion table NN} and~\ref{tab:Heuristics}.

\section{Discussion and Conclusion}
Our results reveal a clear information hierarchy within CoT traces. Lexical uncertainty hints dominate: tokens such as \textit{guess}, \textit{stuck}, and \textit{hard} reduce accuracy odds by up to 40 \% relative to baseline (Fig.\ref{Dragging Words}).
Intra-CoT sentiment volatility is weaker but complementary; a modest upward sentiment shift ($\Delta=0.1$) coincides with peak accuracy on Omni-MATH, yet sentiment is uninformative on HLE
(Fig.\ref{fig:emotion-hedging-combined}). On the moderate difficulty Omni MATH benchmark we find a negative correlation between CoT length and answer accuracy, whereas no such link appears on the more demanding HLE set. This pattern matches earlier work reporting accuracy declines with longer CoTs on Omni MATH \citep{ballon2025relationship} and the absence of this effect on highly challenging benchmarks \citep{illusion-of-thinking,opus2025comment}.

This asymmetry can be exploited to obtain a practical heuristic: filter outputs whose CoT contains strong uncertainty words regardless of length or self-reported confidence. We found that even a small lexicon of “harmful” words drives performance on par with, or better than, more elaborate confidence‐based schemes. Concretely, training a binary classifier on CoT length, emotion volatility, hedging rate plus the top 25 non‐lemmatized “bad” words yields Matthews correlation coefficients (MCCs) of 0.229 when training and testing on HLE (and 0.239 cross‐evaluated on Omni-Math), and 0.354 on Omni-Math (0.263 cross‐evaluated on HLE). By contrast, simply thresholding self‐reported confidence gives MCCs of only 0.085 (HLE) and 0.065 (Omni-Math). Strikingly, a lightweight rule, i.e. mark an answer wrong if any of the top 5 harmful words appears in its CoT, otherwise classify as correct, achieves MCC = 0.215 on HLE and 0.305 on Omni-Math. This suggests that a handful of carefully selected lexical cues can serve as a highly practical post‐hoc filter. A further advantage of this technique is that it treats the CoT as plain text: no weight access, no second inference pass is necessary.

The contrast between DeepSeek-R1 and Claude 3.7 Sonnet illuminates the role of RLHF stylistics. Claude tends to generate longer, optimism-biased CoTs yet shares the same lexical uncertainty fingerprint as R1. This suggests that affective style might be an alignment artefact, whereas the full chain may not mirror the true multi-step search in latent space; the presence of hedging words does correlate with lower internal confidence \citep{chen2025reasoning}.

\subsection*{Limitations and future directions}
In this paper, two English benchmarks were studied, so our findings might not hold up in languages with richer morphology and different hedging forms. Furthermore, our hedging list is fixed and English-focused, meaning we miss any new or non-English uncertainty phrases. The sentiment model we used (OpenAI o3-mini) brings its own calibration quirks. Using multiple raters or a human-scored baseline could smooth out existing noise. Additionally, our HLE grading depends on LLM–human agreement, which implies that the mislabels could hide subtle effects. Finally, correlation is not causation: those uncertainty markers could simply be more common around tougher, hidden problems in general.

Key avenues for future work include: (i) multilingual replication with adaptive hedging lexicons; (ii) adversarial prompting to force models to suppress hedging and test causal impact on accuracy; (iii) To understand RLHF's impact on uncertainty expression, future work could track changes in hedging frequency and harmful token usage between SFT and RLHF checkpoints. This includes testing whether RLHF amplifies or suppresses the lexical signals that predict incorrect reasoning.

\clearpage

\section*{Acknowledgements}
Andres Algaba acknowledges a fellowship from the Research Foundation Flanders under Grant No.1286924N. Vincent Ginis acknowledges support from Research Foundation Flanders under Grant No.G032822N and G0K9322N.

\bibliographystyle{unsrt} 

\clearpage

\setcounter{figure}{0}
\renewcommand{\thefigure}{A\arabic{figure}}
\setcounter{table}{0}
\renewcommand{\thetable}{A\arabic{table}}

\appendix
\section{Accuracy Breakdown}
\label{Accuracy Breakdown}

\begin{table}[ht]
\caption{Accuracy (with count) for each model by HLE category, Omni-MATH tier, and Omni-MATH domain. For questions in Omni-MATH that span multiple domains, each domain instance is counted. Tiers are defined according to~\citep{ballon2025relationship}. For both DeepSeek-R1 and Claude 3.7 Sonnet, the highest accuracy in each benchmark is shown in \textbf{bold}, and the second-highest is \underline{underlined}. While both models excel in similar mathematical sub-disciplines on Omni-MATH, their areas of strength differ on HLE.}
\centering
\resizebox{0.97\textwidth}{!}{%
\begin{tabular}{lllcc}
\toprule
\textbf{Dataset} & \textbf{Subset} & \textbf{Category/Tier/Domain} & 
\textbf{DeepSeek-R1 (n)} & 
\textbf{Claude 3.7 Sonnet (n)} \\
\midrule

\multirow{8}{*}{HLE} & \multirow{8}{*}{Category}
    & Engineering                & \textbf{15\%} (34)   & 6\% (33) \\
&  & Math                       & \underline{9\%} (793)  & \textbf{10\%} (804) \\
&  & Biology/Medicine           & 7\% (68)   & 6\% (66) \\
&  & Computer Science/AI        & 6\% (138)  & 6\% (145) \\
&  & Humanities/Social Science  & 5\% (95)   & \underline{8\%} (90) \\
&  & Physics                    & 5\% (141)  & 6\% (139) \\
&  & Other                      & 3\% (95)   & 2\% (98) \\
&  & Chemistry                  & 2\% (63)   & 5\% (63) \\
\addlinespace

\multirow{4}{*}{Omni-MATH} & \multirow{4}{*}{Tier}
    & Tier 1   & 82\% (1445) & 80\% (1445) \\
&  & Tier 2   & 70\% (1304) & 64\% (1304) \\
&  & Tier 3   & 66\% (649)  & 61\% (649) \\
&  & Tier 4   & 67\% (1030) & 66\% (1030) \\
\addlinespace

\multirow{8}{*}{Omni-MATH} & \multirow{8}{*}{Domain}
    & Algebra               & \textbf{78\%} (2139)  & \textbf{76\%} (2139) \\
&  & Precalculus           & \underline{76}\% (87)    & \underline{75\%} (87) \\
&  & Other                 & 75\% (4)     & \underline{75\%} (4) \\
&  & Number Theory         & 74\% (916)   & 74\% (916) \\
&  & Calculus              & 73\% (128)   & 66\% (128) \\
&  & Applied Mathematics   & 71\% (805)   & 65\% (805) \\
&  & Geometry              & 65\% (1018)  & 60\% (1018) \\
&  & Discrete Mathematics  & 58\% (888)   & 55\% (888) \\

\bottomrule\\
\end{tabular}%
}
\label{accuracy-breakdown}
\end{table}

\clearpage
\section{Booster Words}
\label{Booster Words}
\begin{figure}[ht]
    \centering
    \includegraphics[width=1\linewidth]{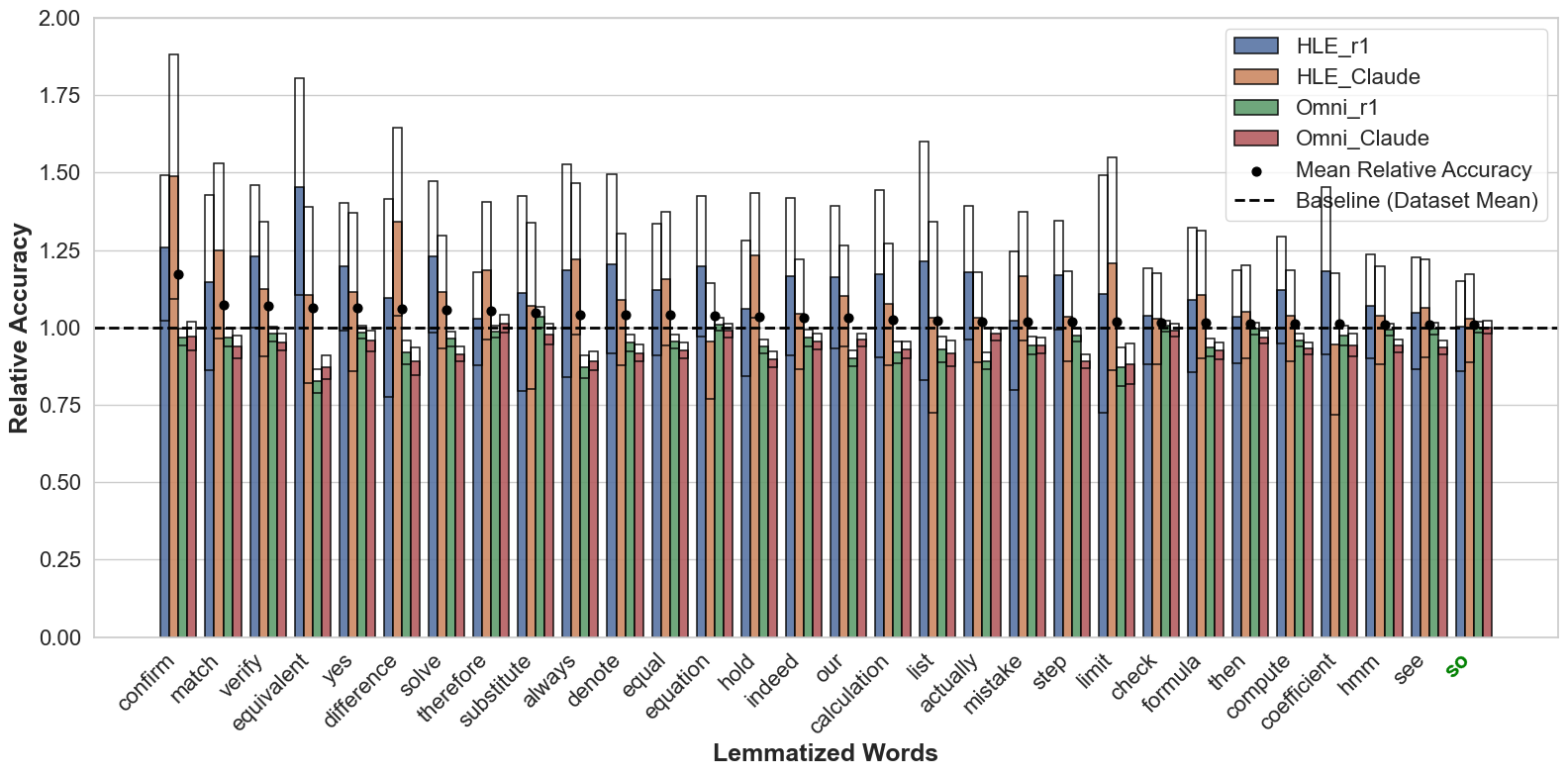}
    \caption{Relative accuracy of the 30 most beneficial lemmatized words across datasets and models. Words characteristic of mathematical reasoning (e.g., solve, therefore, equation) are most associated with higher accuracy. Only ``so'' (in green) appears universally positive, but this is likely due to chance.}
    \label{fig:Booster Words}
\end{figure}

\clearpage
\section{ROC curves}
\label{ROC}

\begin{figure}[htbp]
    \centering
    \begin{tabular}{cc}
        \begin{subfigure}{0.47\textwidth}
            \centering
            \includegraphics[width=\linewidth]{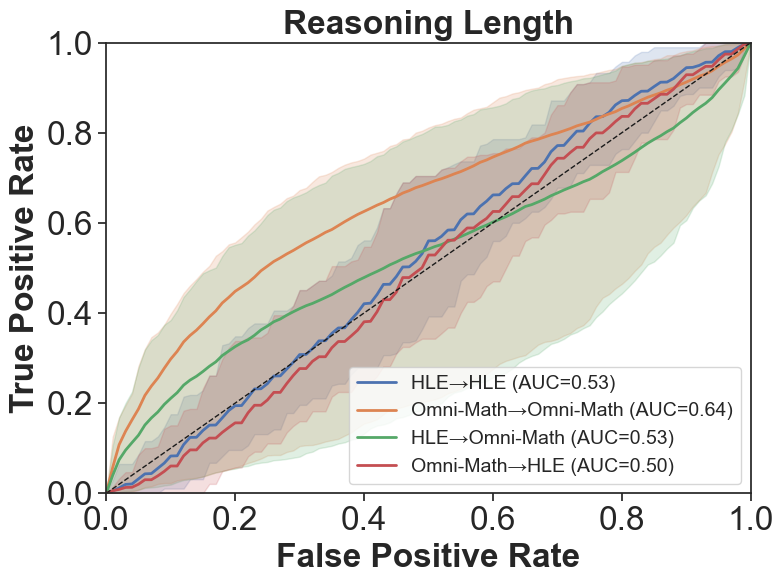}
        \end{subfigure}
        &
        \begin{subfigure}{0.47\textwidth}
            \centering
            \includegraphics[width=\linewidth]{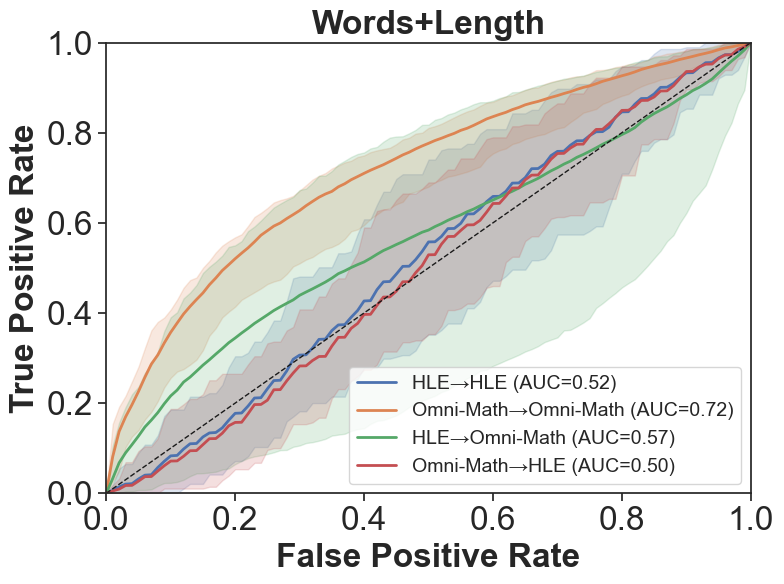}
        \end{subfigure}
        \\
        \begin{subfigure}{0.47\textwidth}
            \centering
            \includegraphics[width=\linewidth]{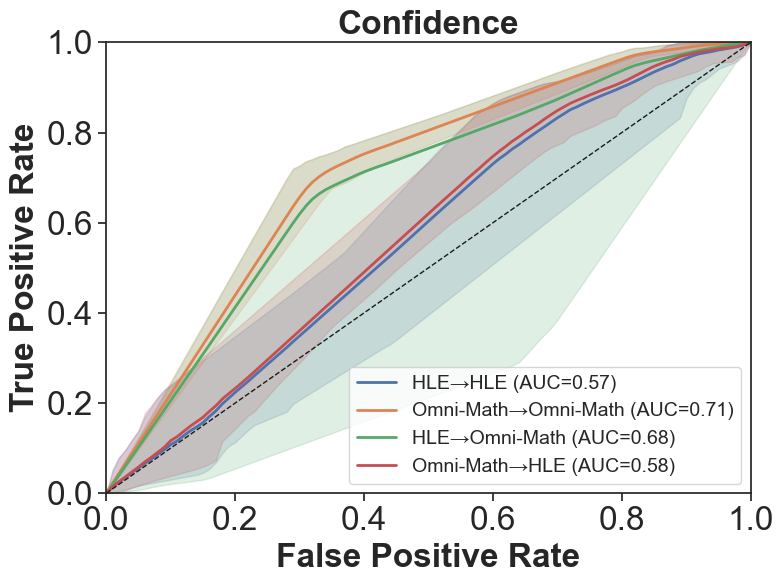}
        \end{subfigure}
        &
        \begin{subfigure}{0.47\textwidth}
            \centering
            \includegraphics[width=\linewidth]{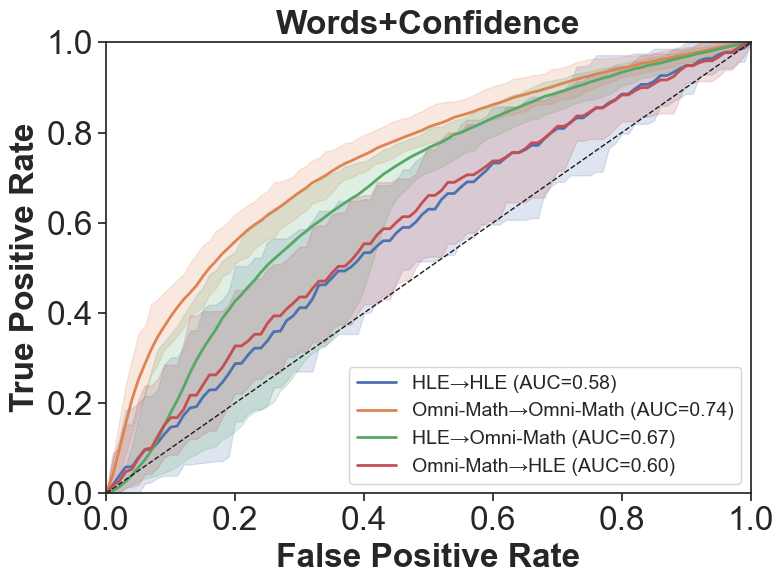}
        \end{subfigure}
        \\
        \begin{subfigure}{0.47\textwidth}
            \centering
            \includegraphics[width=\linewidth]{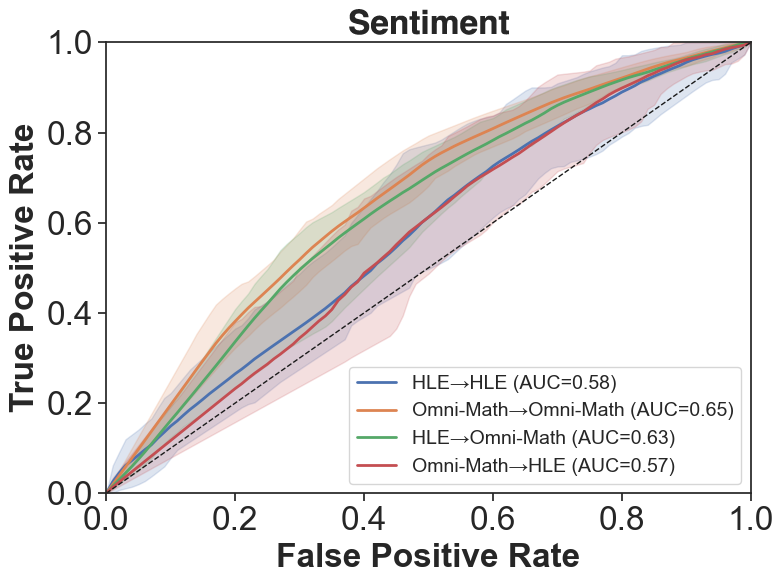}
        \end{subfigure}
        &
        \begin{subfigure}{0.47\textwidth}
            \centering
            \includegraphics[width=\linewidth]{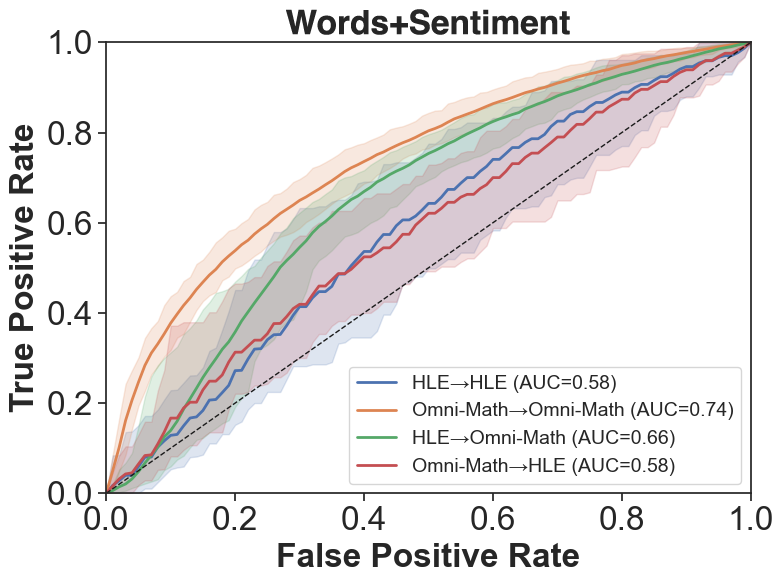}
        \end{subfigure}
    \end{tabular}
\end{figure}

\begin{figure}[htbp]
    \centering
    \begin{tabular}{cc}
        \begin{subfigure}{0.47\textwidth}
            \centering
            \includegraphics[width=\linewidth]{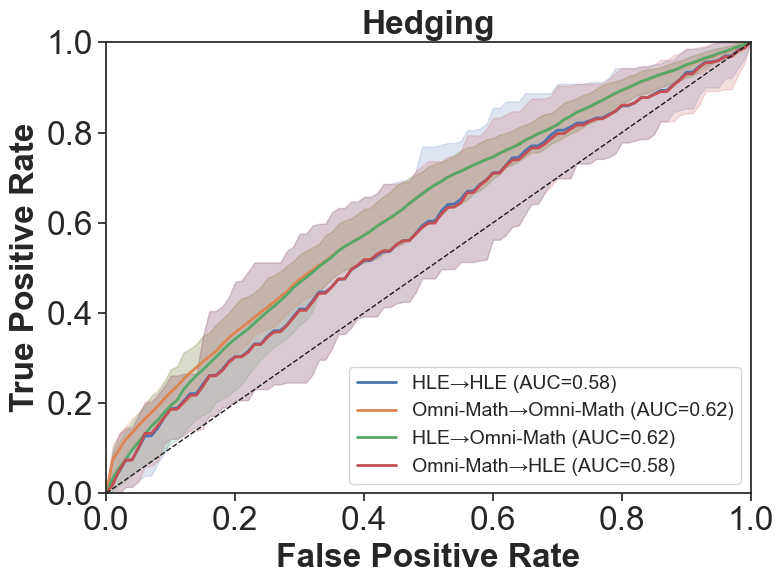}
        \end{subfigure}
        &
        \begin{subfigure}{0.47\textwidth}
            \centering
            \includegraphics[width=\linewidth]{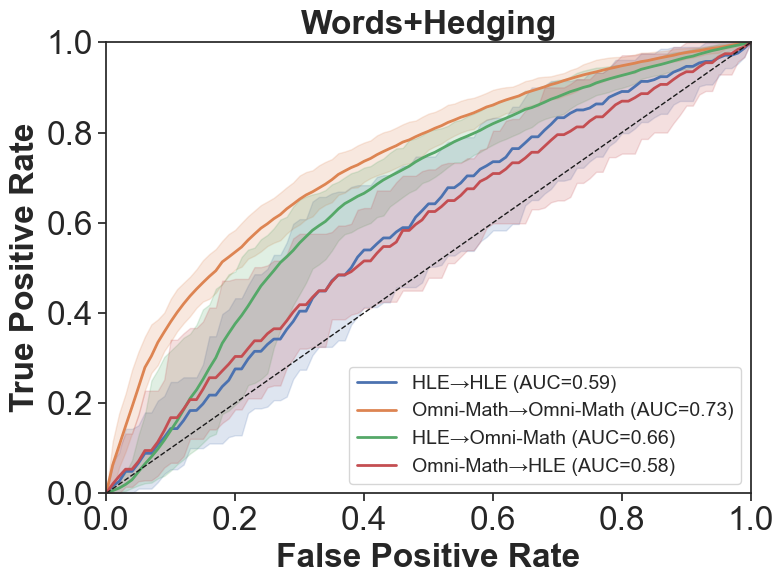}
        \end{subfigure}
        \\
        \begin{subfigure}{0.47\textwidth}
            \centering
            \includegraphics[width=\linewidth]{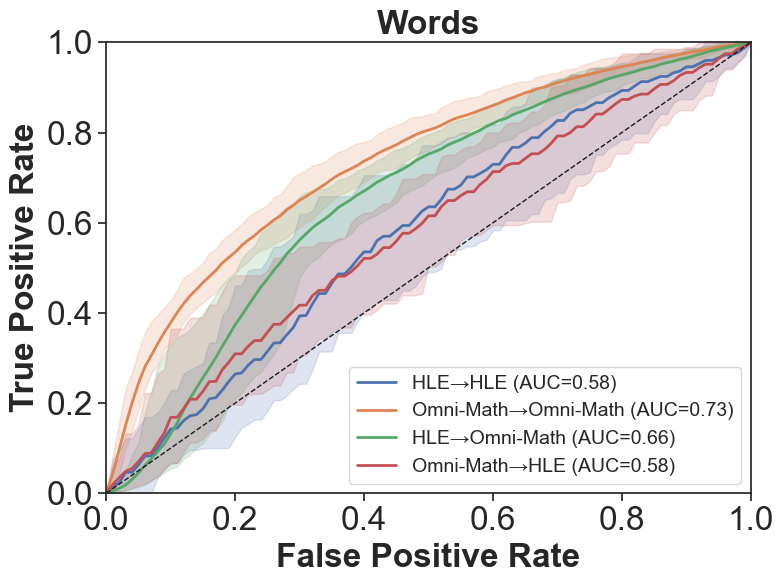}
        \end{subfigure}
        &
        \begin{subfigure}{0.47\textwidth}
            \centering
            \includegraphics[width=\linewidth]{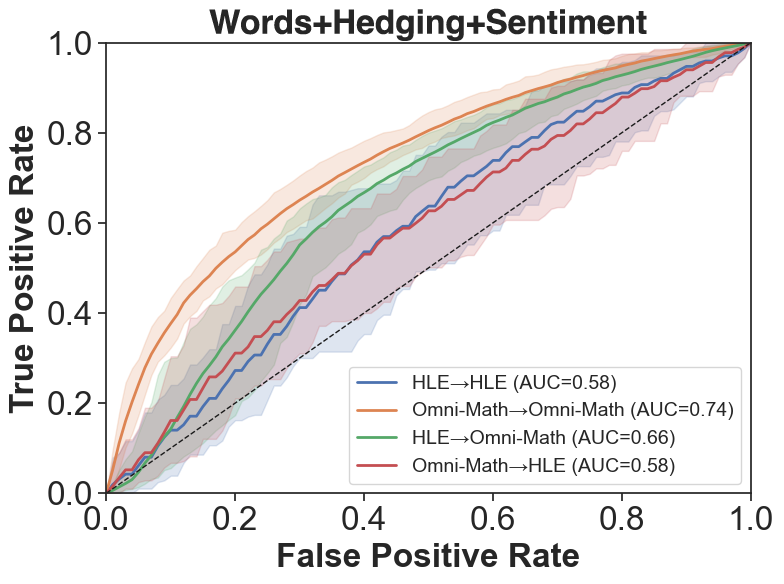}
        \end{subfigure}
        \\
        \begin{subfigure}{0.47\textwidth}
            \centering
            \includegraphics[width=\linewidth]{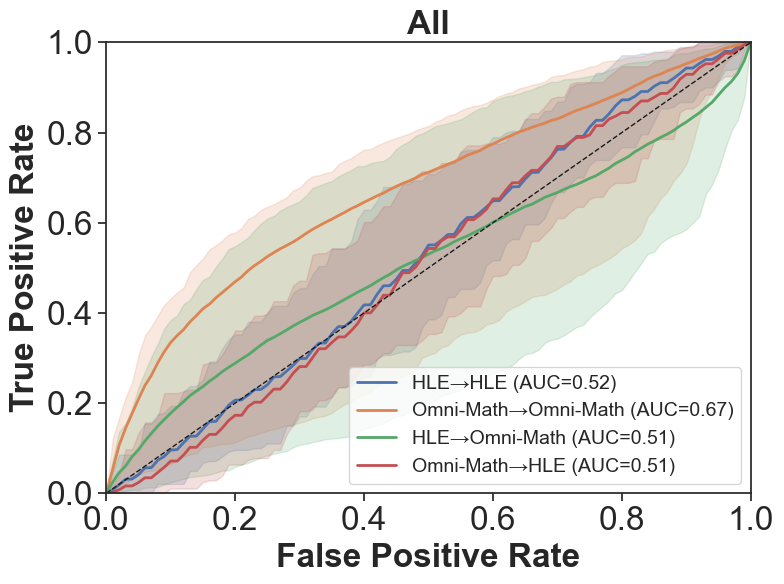}
        \end{subfigure}
        &
        \begin{subfigure}{0.47\textwidth}
            \centering
            \includegraphics[width=\linewidth]{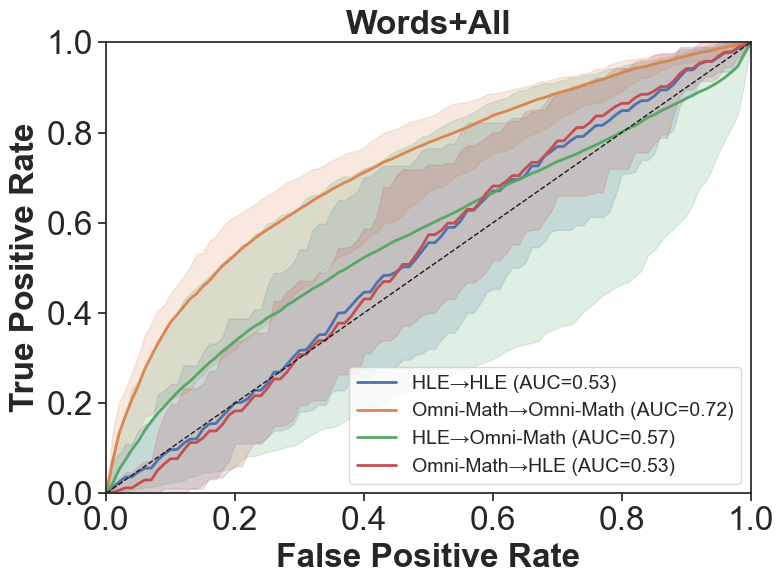}
        \end{subfigure}
    \end{tabular}
    \caption{
ROC curves across 12 experimental settings evaluating neural network classifiers for answer correctness. Each curve displays the mean true/false positive rates and $95\%$ confidence interval (shaded region) across 30 random seeds. The models are trained and evaluated on balanced data for robust comparison. Area under the curve (AUC) values are reported in the legend. Models that use ``harmful" words and abstain from using reasoning length perform the best and most consistently.}
\end{figure}

\clearpage
\section{Consistently Harmful Words Used in Training}
\begin{table}[ht]
\caption{List of 25 non-lemmatized words most strongly and consistently associated with reduced accuracy (``harmful'' words) across benchmarks.}
\centering
\resizebox{0.75\textwidth}{!}{%
\begin{tabular}{lllll}
\toprule
\multicolumn{5}{c}{\textbf{Harmful Words for Accuracy}} \\
\midrule
complexity   & guess      & stuck        & hard         & involved \\
positions    & involves   & involve      & information  & comes    \\
likely       & probably   & constraints  & called       & depend   \\
missing      & often      & possibly     & combination  & based    \\
four         & require    & structure    & per          & requires \\
\bottomrule\\
\end{tabular}%
}
\label{tab:harmful-words}
\end{table}

\clearpage
\section{All Consistently Harmful Words}
\label{appendix: All Consistently Harmful Words}
\begin{table}[ht]
\caption{List of all consistently ``harmful'' words and their 95\% confidence interval for which the offset is given in (brackets). The 25 on average most ``harmful'' words were used as features for training the accuracy predicting neural network. As a consequence of rounding, some words have a relative accuracy equal to one. }
\centering
\resizebox{\textwidth}{!}{%
\begin{tabular}{lccccc}
\toprule
\textbf{Lemmatized Word} & \textbf{Mean Relative Accuracy} & \textbf{HLE-DeepSeek-R1} & \textbf{HLE-Claude}  & \textbf{Omni-DeepSeek-R1} & \textbf{Omni-Claude} \\ \hline
complexity      & 0.59 (0.09)            & 0.61 (0.29)     & 0.54 (0.24) & 0.50 (0.07)      & 0.71 (0.08) \\
guess           & 0.67 (0.17)            & 0.69 (0.29)     & 0.76 (0.23) & 0.42 (0.06)      & 0.80 (0.05) \\
stuck           & 0.70 (0.13)            & 0.71 (0.29)     & 0.81 (0.32) & 0.51 (0.06)      & 0.78 (0.06) \\
hard            & 0.72 (0.12)            & 0.58 (0.25)     & 0.85 (0.24) & 0.65 (0.06)      & 0.79 (0.04) \\
involve         & 0.74 (0.06)            & 0.77 (0.16)     & 0.65 (0.14) & 0.76 (0.03)      & 0.77 (0.04) \\
information     & 0.75 (0.13)            & 0.57 (0.24)     & 0.88 (0.21) & 0.75 (0.06)      & 0.82 (0.05) \\
come            & 0.76 (0.03)            & 0.76 (0.31)     & 0.77 (0.28) & 0.80 (0.07)      & 0.72 (0.07) \\
likely          & 0.77 (0.11)            & 0.66 (0.18)     & 0.92 (0.19) & 0.73 (0.04)      & 0.76 (0.04) \\
probably        & 0.77 (0.03)            & 0.81 (0.25)     & 0.76 (0.22) & 0.78 (0.05)      & 0.75 (0.05) \\
position        & 0.79 (0.04)            & 0.74 (0.20)     & 0.82 (0.18) & 0.80 (0.03)      & 0.79 (0.03) \\
constraint      & 0.79 (0.09)            & 0.68 (0.26)     & 0.82 (0.21) & 0.76 (0.04)      & 0.89 (0.03) \\
call            & 0.79 (0.03)            & 0.75 (0.32)     & 0.80 (0.26) & 0.80 (0.06)      & 0.83 (0.07) \\
miss            & 0.80 (0.10)            & 0.86 (0.27)     & 0.88 (0.20) & 0.67 (0.05)      & 0.77 (0.04) \\
often           & 0.80 (0.03)            & 0.83 (0.26)     & 0.80 (0.22) & 0.75 (0.06)      & 0.81 (0.08) \\
possibly        & 0.80 (0.05)            & 0.79 (0.29)     & 0.87 (0.24) & 0.76 (0.06)      & 0.79 (0.05) \\
combination     & 0.80 (0.09)            & 0.78 (0.27)     & 0.70 (0.27) & 0.90 (0.04)      & 0.84 (0.05) \\
base            & 0.81 (0.09)            & 0.76 (0.20)     & 0.93 (0.16) & 0.72 (0.04)      & 0.83 (0.03) \\
structure       & 0.81 (0.07)            & 0.78 (0.20)     & 0.92 (0.20) & 0.78 (0.04)      & 0.77 (0.04) \\
per             & 0.81 (0.07)            & 0.90 (0.24)     & 0.73 (0.27) & 0.81 (0.05)      & 0.81 (0.07) \\
depend          & 0.82 (0.03)            & 0.84 (0.14)     & 0.86 (0.14) & 0.79 (0.03)      & 0.79 (0.03) \\
complex         & 0.82 (0.09)            & 0.76 (0.22)     & 0.94 (0.18) & 0.74 (0.04)      & 0.84 (0.03) \\
etc             & 0.82 (0.07)            & 0.88 (0.21)     & 0.72 (0.23) & 0.86 (0.03)      & 0.83 (0.04) \\
very            & 0.82 (0.07)            & 0.76 (0.23)     & 0.90 (0.17) & 0.78 (0.04)      & 0.86 (0.03) \\
beyond          & 0.82 (0.04)            & 0.88 (0.33)     & 0.79 (0.32) & 0.81 (0.05)      & 0.81 (0.06) \\
help            & 0.82 (0.05)            & 0.87 (0.26)     & 0.75 (0.27) & 0.87 (0.03)      & 0.82 (0.05) \\
direction       & 0.83 (0.05)            & 0.87 (0.31)     & 0.87 (0.25) & 0.77 (0.05)      & 0.79 (0.04) \\
possibility     & 0.83 (0.10)            & 0.70 (0.29)     & 0.81 (0.24) & 0.91 (0.06)      & 0.89 (0.05) \\
else            & 0.83 (0.05)            & 0.88 (0.27)     & 0.79 (0.26) & 0.85 (0.04)      & 0.80 (0.05) \\
actual          & 0.83 (0.10)            & 0.94 (0.25)     & 0.89 (0.24) & 0.80 (0.04)      & 0.71 (0.05) \\
mention         & 0.83 (0.12)            & 0.89 (0.32)     & 0.97 (0.24) & 0.72 (0.07)      & 0.75 (0.05) \\
group           & 0.84 (0.12)            & 0.66 (0.26)     & 0.94 (0.30) & 0.88 (0.06)      & 0.86 (0.06) \\
without         & 0.84 (0.06)            & 0.85 (0.18)     & 0.92 (0.17) & 0.78 (0.03)      & 0.80 (0.03) \\
require         & 0.84 (0.06)            & 0.81 (0.12)     & 0.92 (0.16) & 0.79 (0.02)      & 0.83 (0.03) \\
far             & 0.84 (0.06)            & 0.80 (0.31)     & 0.90 (0.18) & 0.77 (0.05)      & 0.88 (0.03) \\
additional      & 0.84 (0.06)            & 0.88 (0.32)     & 0.90 (0.23) & 0.75 (0.06)      & 0.83 (0.04) \\
space           & 0.84 (0.10)            & 0.98 (0.29)     & 0.86 (0.23) & 0.78 (0.07)      & 0.75 (0.07) \\

\bottomrule\\
\multicolumn{6}{c}{\textbf{Table continues on the next two pages}}    
\end{tabular}%
}
\label{tab:Consistently Harmful Words 1}
\end{table}

\begin{table}[ht]
\centering
\resizebox{\textwidth}{!}{%
\begin{tabular}{llllll}
\toprule
\textbf{Lemmatized Word} & \textbf{Mean Relative Accuracy} & \textbf{HLE-DeepSeek-R1} & \textbf{HLE-Claude}  & \textbf{Omni-DeepSeek-R1} & \textbf{Omni-Claude} \\ \hline
system          & 0.84 (0.02)            & 0.85 (0.27)     & 0.81 (0.23) & 0.85 (0.04)      & 0.86 (0.04) \\
too             & 0.84 (0.08)            & 0.75 (0.18)     & 0.95 (0.19) & 0.81 (0.03)      & 0.85 (0.03) \\
certain         & 0.85 (0.07)            & 0.91 (0.19)     & 0.90 (0.19) & 0.77 (0.03)      & 0.81 (0.04)\\
lead            & 0.85 (0.06)            & 0.78 (0.13)     & 0.91 (0.15) & 0.86 (0.02)      & 0.85 (0.03) \\
part            & 0.86 (0.04)            & 0.82 (0.15)     & 0.89 (0.16) & 0.88 (0.03)      & 0.82 (0.03) \\
unless          & 0.86 (0.02)            & 0.87 (0.21)     & 0.83 (0.28) & 0.85 (0.03)      & 0.87 (0.05) \\
specific        & 0.86 (0.03)            & 0.88 (0.19)     & 0.88 (0.15) & 0.81 (0.03)      & 0.86 (0.03) \\
proceed         & 0.86 (0.12)            & 1.00 (0.26)     & 0.88 (0.32) & 0.70 (0.05)      & 0.86 (0.06) \\
change          & 0.86 (0.08)            & 0.88 (0.22)     & 0.96 (0.18) & 0.78 (0.04)      & 0.82 (0.03) \\
common          & 0.86 (0.09)            & 0.94 (0.26)     & 0.74 (0.21) & 0.86 (0.04)      & 0.91 (0.04) \\
pattern         & 0.86 (0.06)            & 0.78 (0.30)     & 0.92 (0.22) & 0.85 (0.03)      & 0.89 (0.03) \\
high            & 0.86 (0.09)            & 0.94 (0.28)     & 0.94 (0.28) & 0.80 (0.05)      & 0.77 (0.07) \\
suggest         & 0.86 (0.09)            & 0.81 (0.20)     & 1.00 (0.18) & 0.79 (0.03)      & 0.86 (0.04) \\
angle           & 0.86 (0.05)            & 0.90 (0.25)     & 0.90 (0.19) & 0.81 (0.04)      & 0.84 (0.03) \\
confuse         & 0.86 (0.10)            & 0.92 (0.25)     & 0.98 (0.25) & 0.77 (0.05)      & 0.79 (0.05) \\
assumption      & 0.87 (0.07)            & 0.92 (0.26)     & 0.90 (0.27) & 0.77 (0.04)      & 0.90 (0.05) \\
analysis        & 0.87 (0.06)            & 0.94 (0.33)     & 0.87 (0.21) & 0.79 (0.05)      & 0.88 (0.04) \\
complicated     & 0.87 (0.03)            & 0.92 (0.19)     & 0.84 (0.22) & 0.87 (0.03)      & 0.86 (0.03) \\
similar         & 0.87 (0.08)            & 0.97 (0.20)     & 0.91 (0.21) & 0.79 (0.03)      & 0.82 (0.04) \\
multiple        & 0.88 (0.04)            & 0.91 (0.23)     & 0.83 (0.20) & 0.86 (0.03)      & 0.90 (0.03) \\
much            & 0.88 (0.07)            & 0.90 (0.26)     & 0.96 (0.20) & 0.82 (0.04)      & 0.83 (0.03)\\
specifically    & 0.88 (0.04)            & 0.93 (0.28)     & 0.89 (0.17) & 0.85 (0.05)      & 0.85 (0.03) \\
relate          & 0.88 (0.07)            & 0.92 (0.17)     & 0.96 (0.17) & 0.84 (0.03)      & 0.80 (0.03)\\
next            & 0.88 (0.05)            & 0.91 (0.23)     & 0.81 (0.22) & 0.93 (0.03)      & 0.87 (0.03) \\
count           & 0.88 (0.04)            & 0.89 (0.28)     & 0.94 (0.26) & 0.87 (0.04)      & 0.84 (0.04) \\
new             & 0.89 (0.07)            & 0.90 (0.30)     & 0.98 (0.23) & 0.83 (0.04)      & 0.83 (0.04) \\
relation        & 0.89 (0.03)            & 0.85 (0.31)     & 0.93 (0.25) & 0.87 (0.05)      & 0.89 (0.04) \\
assume          & 0.89 (0.07)            & 0.95 (0.15)     & 0.93 (0.14) & 0.80 (0.03)      & 0.87 (0.02) \\
simple          & 0.89 (0.07)            & 0.97 (0.25)     & 0.92 (0.14) & 0.81 (0.04)      & 0.84 (0.02) \\
third           & 0.89 (0.07)            & 0.80 (0.25)     & 0.97 (0.25) & 0.90 (0.03)      & 0.88 (0.03) \\
look            & 0.89 (0.07)            & 0.83 (0.16)     & 0.98 (0.12) & 0.84 (0.03)      & 0.90 (0.02) \\
thing           & 0.89 (0.06)            & 0.96 (0.26)     & 0.92 (0.22) & 0.82 (0.04)      & 0.86 (0.04) \\
within          & 0.90 (0.08)            & 0.95 (0.30)     & 0.97 (0.24) & 0.82 (0.04)      & 0.83 (0.05) \\
correspond      & 0.90 (0.07)            & 0.94 (0.17)     & 0.97 (0.15) & 0.82 (0.03)      & 0.85 (0.03) \\
lower           & 0.90 (0.10)            & 0.99 (0.26)     & 0.98 (0.29) & 0.82 (0.04)      & 0.80 (0.06) \\
valid           & 0.90 (0.07)            & 0.80 (0.27)     & 0.97 (0.23) & 0.90 (0.03)      & 0.91 (0.03) \\
\bottomrule\\
\multicolumn{6}{c}{\textbf{Table continues on the next page}}  
\end{tabular}%
}
\label{tab: Consistently Harmful Words 2}
\end{table}
\begin{table}[ht]
\centering
\resizebox{\textwidth}{!}{%
\begin{tabular}{llllll}
\toprule
\textbf{Lemmatized Word} & \textbf{Mean Relative Accuracy} & \textbf{HLE-DeepSeek-R1} & \textbf{HLE-Claude}  & \textbf{Omni-DeepSeek-R1} & \textbf{Omni-Claude} \\ \hline
single          & 0.90 (0.10)            & 0.99 (0.23)     & 0.98 (0.21) & 0.81 (0.04)      & 0.81 (0.04) \\
pair            & 0.90 (0.03)            & 0.88 (0.23)     & 0.95 (0.21) & 0.89 (0.03)      & 0.88 (0.03) \\
form            & 0.90 (0.05)            & 0.85 (0.20)     & 0.97 (0.17) & 0.88 (0.03)      & 0.92 (0.03) \\
something       & 0.90 (0.05)            & 0.95 (0.19)     & 0.94 (0.17) & 0.88 (0.03)      & 0.85 (0.03) \\
itself          & 0.91 (0.05)            & 0.94 (0.25)     & 0.95 (0.22) & 0.86 (0.04)      & 0.86 (0.04) \\
those           & 0.91 (0.06)            & 0.97 (0.20)     & 0.91 (0.20) & 0.91 (0.03)      & 0.84 (0.03) \\
distinct        & 0.91 (0.08)            & 0.99 (0.35)     & 0.96 (0.27) & 0.86 (0.04)      & 0.83 (0.04)\\
general         & 0.91 (0.03)            & 0.94 (0.24)     & 0.93 (0.18) & 0.89 (0.04)      & 0.89 (0.03) \\
don             & 0.91 (0.03)            & 0.95 (0.20)     & 0.92 (0.15) & 0.89 (0.03)      & 0.89 (0.02) \\
integer         & 0.91 (0.05)            & 0.86 (0.21)     & 0.89 (0.19) & 0.93 (0.02)      & 0.97 (0.02) \\
why             & 0.91 (0.09)            & 0.99 (0.36)     & 1.00 (0.24) & 0.86 (0.05)      & 0.81 (0.04) \\
even            & 0.91 (0.03)            & 0.95 (0.18)     & 0.90 (0.16) & 0.91 (0.02)      & 0.89 (0.03) \\
could           & 0.91 (0.05)            & 0.93 (0.17)     & 0.98 (0.16) & 0.87 (0.03)      & 0.88 (0.03) \\
their           & 0.91 (0.06)            & 0.99 (0.18)     & 0.89 (0.18) & 0.91 (0.03)      & 0.86 (0.03) \\
still           & 0.92 (0.06)            & 0.91 (0.18)     & 0.99 (0.15) & 0.86 (0.03)      & 0.90 (0.02) \\
add             & 0.92 (0.04)            & 0.91 (0.19)     & 0.96 (0.20) & 0.95 (0.02)      & 0.86 (0.03) \\
linear          & 0.92 (0.05)            & 0.99 (0.28)     & 0.89 (0.27) & 0.90 (0.04)      & 0.90 (0.05) \\
factor          & 0.92 (0.08)            & 0.95 (0.19)     & 0.80 (0.18) & 0.98 (0.02)      & 0.96 (0.03) \\
consider        & 0.92 (0.05)            & 0.97 (0.12)     & 0.96 (0.11) & 0.88 (0.02)      & 0.88 (0.02) \\
figure          & 0.93 (0.07)            & 0.94 (0.18)     & 0.99 (0.25) & 0.94 (0.03)      & 0.83 (0.04) \\
through         & 0.93 (0.06)            & 0.97 (0.23)     & 0.99 (0.17) & 0.86 (0.03)      & 0.88 (0.03) \\
such            & 0.93 (0.02)            & 0.90 (0.16)     & 0.95 (0.16) & 0.92 (0.02)      & 0.93 (0.02) \\
question        & 0.93 (0.07)            & 0.98 (0.17)     & 0.99 (0.16) & 0.90 (0.03)      & 0.85 (0.03) \\
however         & 0.93 (0.07)            & 0.99 (0.15)     & 0.99 (0.15) & 0.88 (0.02)      & 0.87 (0.03) \\
start           & 0.93 (0.05)            & 0.95 (0.14)     & 0.99 (0.13) & 0.93 (0.02)      & 0.86 (0.02) \\
like            & 0.93 (0.03)            & 0.97 (0.16)     & 0.95 (0.15) & 0.91 (0.02)      & 0.90 (0.02) \\
work            & 0.93 (0.03)            & 0.90 (0.18)     & 0.97 (0.15) & 0.92 (0.02)      & 0.94 (0.02) \\
positive        & 0.94 (0.03)            & 0.93 (0.25)     & 0.89 (0.22) & 0.96 (0.03)      & 0.96 (0.03) \\
express         & 0.94 (0.03)            & 0.96 (0.20)     & 0.97 (0.19) & 0.90 (0.03)      & 0.92 (0.03) \\
example         & 0.94 (0.05)            & 0.96 (0.15)     & 1.00 (0.20) & 0.92 (0.02)      & 0.88 (0.03) \\
they            & 0.94 (0.04)            & 0.99 (0.12)     & 0.97 (0.12) & 0.92 (0.02)      & 0.90 (0.02) \\
my              & 0.95 (0.07)            & 0.99 (0.23)     & 1.00 (0.14) & 0.85 (0.04)      & 0.94 (0.02) \\
up              & 0.95 (0.04)            & 0.98 (0.17)     & 0.97 (0.16) & 0.94 (0.02)      & 0.89 (0.03) \\
take            & 0.95 (0.05)            & 0.99 (0.16)     & 1.00 (0.14) & 0.93 (0.02)      & 0.88 (0.02) \\
down            & 0.95 (0.07)            & 0.98 (0.23)     & 1.00 (0.19) & 0.97 (0.03)      & 0.85 (0.03) \\
need            & 0.96 (0.02)            & 0.97 (0.12)     & 0.98 (0.11) & 0.94 (0.02)      & 0.94 (0.02) \\
solution        & 0.96 (0.04)            & 1.00 (0.30)     & 0.99 (0.23) & 0.92 (0.03)      & 0.94 (0.03) \\
on              & 0.96 (0.03)            & 0.99 (0.15)     & 0.98 (0.14) & 0.93 (0.02)      & 0.94 (0.02) \\
an              & 0.96 (0.04)            & 0.99 (0.16)     & 0.99 (0.14) & 0.92 (0.02)      & 0.95 (0.02) \\
be              & 0.97 (0.02)            & 0.98 (0.07)     & 0.99 (0.06) & 0.95 (0.01)      & 0.94 (0.01) \\
not             & 0.97 (0.02)            & 0.98 (0.15)     & 1.00 (0.14) & 0.94 (0.02)      & 0.96 (0.02)\\
\bottomrule\\
\end{tabular}%
}
\label{tab:full lexicon harmfull lemma's}
\end{table}

\begin{table}[ht]
\caption{Confusion matrices and Matthews Correlation Coefficients (MCC) for our best-performing neural network, trained using only the 25 most consistently harmful words as input features. Evaluations were conducted on a balanced test set, with results shown for a single random seed (seed = 2).}
\centering
\begin{tabular}{lrrrr}
\toprule
\textbf{Train$\rightarrow$Test} & \textbf{Actual} & \textbf{Pred. 0} & \textbf{Pred. 1} & \textbf{MCC} \\
\midrule
\multirow{2}{*}{HLE$\rightarrow$HLE}         & 0 & 40  & 30   & \multirow{2}{*}{0.229} \\
                                             & 1 & 24  & 46   &                        \\
\midrule
\multirow{2}{*}{HLE$\rightarrow$Omni-Math}   & 0 & 241 & 276  & \multirow{2}{*}{0.239} \\
                                             & 1 & 123 & 394  &                        \\
\midrule
\multirow{2}{*}{Omni-Math$\rightarrow$Omni-Math} & 0 & 355 & 162  & \multirow{2}{*}{0.354} \\
                                                 & 1 & 172 & 345  &                        \\
\midrule
\multirow{2}{*}{Omni-Math$\rightarrow$HLE}   & 0 & 57  & 13    & \multirow{2}{*}{0.263} \\
                                             & 1 & 40  & 30   &                        \\
\bottomrule\\
\end{tabular}
\label{confusion table NN}
\end{table}

\begin{table}[ht]
\caption{Confusion matrices and Matthews Correlation Coefficients (MCC) for two heuristic methods evaluated on the HLE and Omni-Math datasets using a balanced test set. 
The 5 Harmful Words heuristic predicts failure if any of the five most consistently harmful words appear in the reasoning text. 
The Confidence heuristic uses self-reported confidence to probabilistically predict correctness by flipping coin weighted by its reported confidence level, to assign a grade of either 0 or 1. Our results illustrate that the word-based heuristic consistently outperforms the probabilistic confidence-based method.}
\centering
\begin{tabular}{llrrrr}
\toprule
\textbf{Heuristic} & \textbf{Dataset} & \textbf{Actual} & \textbf{Pred. 0} & \textbf{Pred. 1} & \textbf{MCC} \\
\midrule
\multirow{2}{*}{5 Most Harmful Words} 
  & HLE         & 0 & 46 & 24   & \multirow{2}{*}{0.215} \\
  &             & 1 & 31 & 39   &                         \\
\cmidrule{2-6}
  & Omni-Math   & 0 & 317 & 200  & \multirow{2}{*}{0.305} \\
  &             & 1 & 160 & 357 &                        \\
\midrule
\multirow{2}{*}{Confidence} 
  & HLE         & 0 & 1  & 69  & \multirow{2}{*}{0.085} \\
  &             & 1 & 0  & 70  &                         \\
\cmidrule{2-6}
  & Omni-Math   & 0 & 35 & 482 & \multirow{2}{*}{0.065} \\
  &             & 1 & 20 & 497 &                        \\
\bottomrule \\
\end{tabular}
\label{tab:Heuristics}
\end{table}

\end{document}